\documentclass[10pt,twocolumn,letterpaper]{article}
\pdfoutput=1

\usepackage{cvpr}
\usepackage{times}
\usepackage{epsfig}
\usepackage{graphicx}
\usepackage{amsmath}
\usepackage{amssymb}
\usepackage{amsfonts}
\usepackage{verbatim}
\usepackage{bm}
\usepackage{hhline}
\usepackage[normalem]{ulem}

\usepackage[pagebackref=true,breaklinks=true,letterpaper=true,colorlinks,bookmarks=false]{hyperref}

\newcommand{\overbar}[1]{\mkern 1.5mu\overline{\mkern-1.5mu#1\mkern-1.5mu}\mkern 1.5mu}

\def\ff{\mathbf f}
\def\ll{\mathbf l}

\def\N{\mathcal N}

\def\Re{\mathbb{R}}
\def\Ze{\mathbb{Z}}

\def\btheta{{\bm\theta}}

\def\ind{\textup{ind}}

\makeatletter
\newcommand\footnoteref[1]{\protected@xdef\@thefnmark{\ref{#1}}\@footnotemark}
\makeatother

\cvprfinalcopy 

\def\cvprPaperID{245} 

\ifcvprfinal\fi

\begin{document}

\title{Full Flow: Optical Flow Estimation By Global Optimization over Regular Grids}

\author{Qifeng Chen\\
Stanford University\\
\and
Vladlen Koltun\\
Intel Labs\\
}

\maketitle

\begin{abstract}
We present a global optimization approach to optical flow estimation. The approach optimizes a classical optical flow objective over the full space of mappings between discrete grids. No descriptor matching is used. The highly regular structure of the space of mappings enables optimizations that reduce the computational complexity of the algorithm's inner loop from quadratic to linear and support efficient matching of tens of thousands of nodes to tens of thousands of displacements. We show that one-shot global optimization of a classical Horn-Schunck-type objective over regular grids at a single resolution is sufficient to initialize continuous interpolation and achieve state-of-the-art performance on challenging modern benchmarks.
\end{abstract}

\section{Introduction}
\label{sec:introduction}

Optical flow is a vital source of information for visual perception. Animals use optical flow to track and control self-motion, to estimate the spatial layout of the environment, and to perceive the shape and motion of objects \cite{Srinivasan2011,Warren1995}. In computer vision, optical flow is used for visual odometry, three-dimensional reconstruction, object segmentation and tracking, and recognition.

The classical approach to dense optical flow estimation is to optimize an objective of the form
\begin{equation}
E(\ff) = E_{\text{data}}(\ff) + \lambda E_{\text{reg}}(\ff),
\label{eq:intro}
\end{equation}
where $\ff$ is the estimated flow field, $E_{\text{data}}$ is a data term that penalizes association of visually dissimilar areas, and $E_{\text{reg}}$ is a regularization term that penalizes incoherent motion~\cite{HornSchunck1981}. Traditionally, this objective is optimized by iterative local refinement that maintains and updates a single candidate flow \cite{Sun2014}. This local refinement does not optimize the objective globally over the full space of flows and is prone to local minima.

Global optimization of the flow objective has generally been considered intractable unless significant restrictions are imposed~\cite{Strekalovskiy2014}. Menze et al.~\cite{Menze2015} achieved impressive results with a discrete optimization approach, but had to heuristically prune the space of flows using descriptor matching.

In this work, we develop a global optimization approach that optimizes the classical flow objective (\ref{eq:intro}) over the full space of mappings between discrete grids. Our work demonstrates that a direct application of global optimization over full regular grids has significant benefits. Since the highly regular structure of the space of mappings is preserved, we can employ optimizations that take advantage of this structure to reduce the computational complexity of the algorithm's inner loop. The overall approach is simple and does not involve separately-defined descriptor matching modules: simply optimizing the classical flow objective over full grids is sufficient. We show that this minimalistic approach yields state-of-the-art accuracy on both the Sintel~\cite{Butler2012} and the KITTI 2015 \cite{MenzeGeiger2015} optical flow benchmarks.

\section{Background}
\label{sec:background}

The variational approach to optical flow originates with Horn and Schunck \cite{HornSchunck1981}. This elegant approach posits a clear global objective (\ref{eq:intro}) and produces a dense flow field connecting the two images. Since the space of flows is so large, the variational objective has traditionally been optimized locally. Starting with a simple initialization, the flow is iteratively updated by gradient-based steps \cite{BlackAnandan1996,Brox2004,Sun2014}. Through these iterations, a single candidate flow is maintained. While this local refinement approach can be accurate when displacements are small \cite{Baker2011,Sun2014}, it does not optimize the objective globally over the full space of flows and is prone to local minima.

Recent methods have used descriptor matching and nearest neighbor search to initialize the continuous refinement \cite{BroxMalik2011,Xu2012,Chen2013,Weinzaepfel2013,Revaud2015,Bailer2015}. This more sophisticated initialization is known to significantly improve results in the presence of large displacements. However, the descriptor matching module is trained separately, does not optimize a coherent objective over the provided correspondence sets, and can yield globally suboptimal initializations. We show that state-of-the-art accuracy can be achieved by globally optimizing the classical objective (\ref{eq:intro}), with no separately trained or designed descriptors.

A number of approaches to global optimization for optical flow estimation have been proposed. Steinbr{\"{u}}cker et al.~\cite{Steinbrucker2009} use an alternating scheme to optimize a quadratic relaxation of the global objective. This formulation relies on the assumption that the regularizer is convex. A number of subsequent approaches use functional lifting to map the problem into a higher-dimensional space, where the optimization reduces to estimating a collection of hypersurfaces~\cite{Goldstein2012,Goldluecke2013,Strekalovskiy2014}. These schemes likewise impose certain assumptions on the model, such as requiring the data term or the regularizer to be convex.
In general, these approaches have not been shown to produce state-of-the-art results on modern benchmarks.

Our approach treats objective (\ref{eq:intro}) as a Markov random field and uses discrete optimization techniques. The Markov random field perspective on optical flow estimation dates back to the 80s and discrete optimization techniques have been applied to the problem in different forms since that time \cite{KonradDubois1988,HeitzBouthemy1993}. Glocker et al.~\cite{Glocker2008,Glocker2010} applied MRF optimization to sets of control points in coarse-to-fine schemes. In contrast, we operate on dense grids with large two-dimensional label spaces. A number of works considered a simplified MRF formulation that decomposes the horizontal and vertical components of the flow \cite{Shekhovtsov2008,Lee2010,Zach2014}. In contrast, we demonstrate the feasibility of operating on much larger models with two-dimensional label spaces. Lempitsky et al.~\cite{Lempitsky2010} iteratively improved the estimated flow field by generating proposals and integrating them using the QPBO algorithm. In contrast, we optimize over the full space of mappings between discrete grids. Komodakis et al.~\cite{Komodakis2011} evaluated MRF optimization on optical flow estimation with small displacements. In contrast, we show that global optimization over full two-dimensional label spaces is tractable and yields state-of-the-art performance on challenging large-displacement problems.

Menze et al.~\cite{Menze2015} pruned the space of flows using feature descriptors and optimized an MRF on the pruned label space. In contrast, we argue that operating on the full space is both feasible and desirable. First, we avoid heuristic pruning and the reliance on separately-defined feature descriptors that are not motivated by the flow objective itself. Second, pruning destroys the highly regular structure of the space of mappings. We show that optimization over the full space can be significantly accelerated due to the regularity of the space. In particular, the full regular structure enables the use of highly optimized min-convolution algorithms that reduce the complexity of message passing from quadratic to linear \cite{FelzenszwalbHuttenlocher2012,ChenKoltun2014}.

\section{Model}
\label{sec:model}

\begin{figure}
\centering
\includegraphics[width=0.38\textwidth]{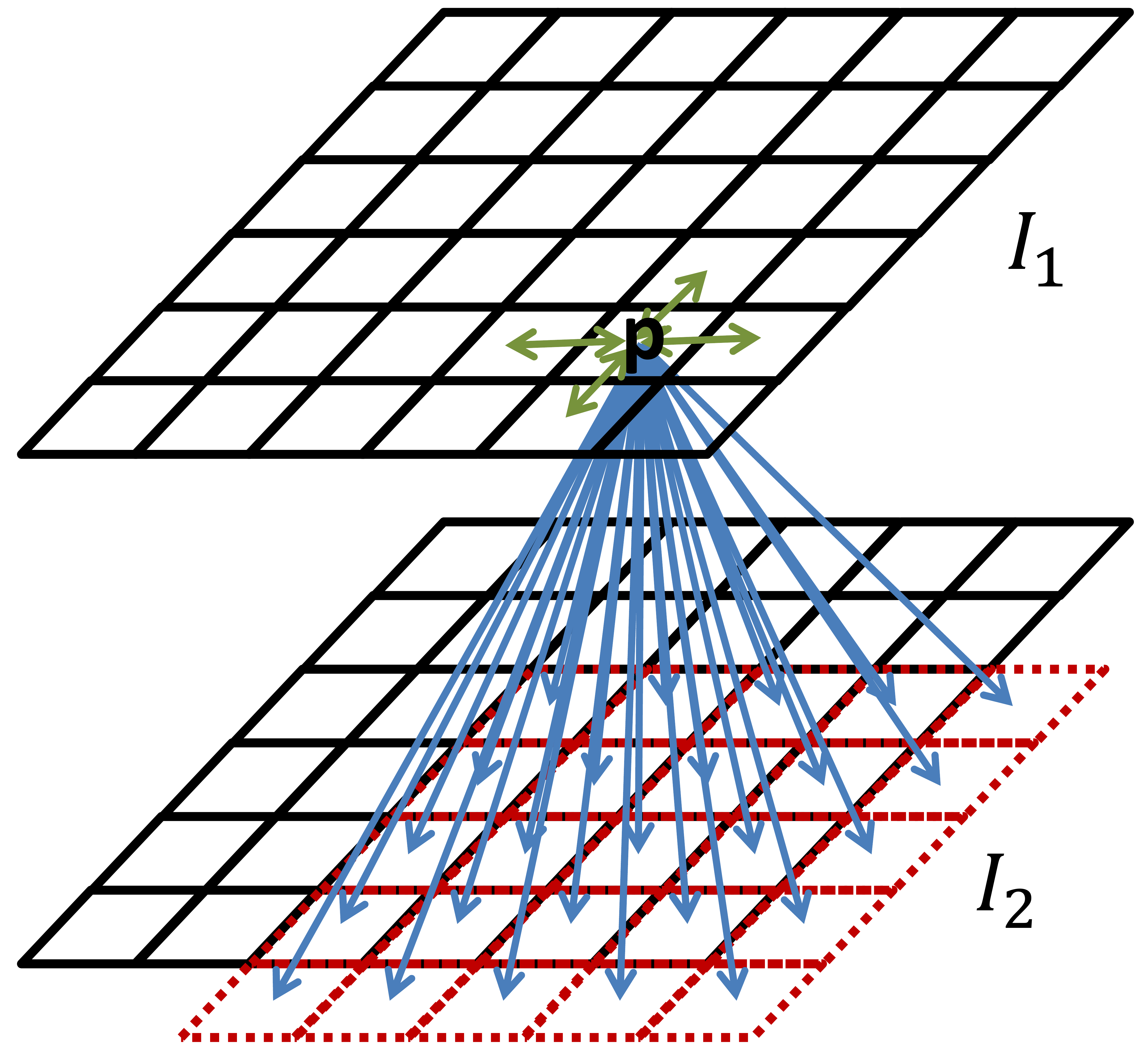}
\vspace{2mm}
\caption{Optical flow over regular grids. Each pixel $p$ in $I_1$ is spatially connected to its four neighbors in $I_1$ and temporally connected to $(2\varsigma +1)^2$ pixels in $I_2$.}
\label{figure:illustration}
\vspace{-2mm}
\end{figure}

Let $I_1, I_2: \Omega \rightarrow \Re^3$ be two color images, where $\Omega \subset \Ze^2$ is the image domain. Let $\ff = (\ff^1,\ff^2): \Omega \rightarrow [-\varsigma,\varsigma]^2$ be a flow field that maps each pixel $p$ in $I_1$ to $(p+f_p)$ in an augmented domain $\overbar{\Omega} \supset \Omega$, which contains $\Omega$ and a large surrounding buffer zone. The buffer zone absorbs pixels that flow out of the visual field. The augmented domain $\overbar{\Omega} \subset \Ze^2$ is the Minkowski sum of $\Omega$ and $[-\varsigma,\varsigma]^2\cap\Ze^2$, where $\varsigma$ is the maximal empirical displacement magnitude. The maximal empirical displacement magnitude is measured by taking the maximal displacement observed in a training set. For example, the maximal displacement magnitude on the KITTI training set \cite{Geiger2013,MenzeGeiger2015} is $242$ pixels. We perform the optimization on $1/3$-resolution images, so $\varsigma = 81$ for the KITTI dataset.

Our objective function is
\begin{eqnarray}
E(\ff) &=& \sum_{p\in I_1}{\rho_D(p,f_p,I_1,I_2)} \nonumber\\
&+& \lambda \sum_{\{p,q\}\in \N}{w_{p,q}\,\rho_S (f_p-f_q)},
\label{eqn:HS}
\end{eqnarray}
where $\N \subset \Omega^2$ is the 4-connected pixel grid. See Figure~\ref{figure:illustration} for illustration. The data term $\rho_D(p,f_p,I_1,I_2)$ penalizes flow fields that connect dissimilar pixels $p$ and $(p+f_p)$. We use truncated normalized cross-correlation \cite{Vogel2013}:
\begin{equation}
  \rho_D(p,f_p,I_1,I_2) = 1 - \max(NCC,0),
  \label{eq:NCC}
\end{equation}
where $NCC$ is the normalized cross-correlation between two patches, one centered at $p$ in $I_1$ and one centered at $(p+f_p)$ in $I_2$, computed in each color channel and averaged. The truncation at zero prevents penalization of negatively correlated patches. If $(p+f_p)$ is in the buffer zone $\overbar{\Omega} \setminus \Omega$, the data term is set to a constant penalty $\zeta$.

Our optimization approach assumes that the regularization term has the following form:
\begin{equation}
  \rho_S(f) = \min \big( \rho(f^1)+\rho(f^2), \tau \big),
  \label{eq:reg-form}
\end{equation}
where $f^1,f^2$ are the two components of vector $f$ and $\rho(\cdot)$ is a penalty function, such as the $L^1$ norm or the Charbonnier penalty. Our formulation and the general solution strategy can accomodate non-convex functions $\rho$, such as the Lorentzian and the generalized Charbonnier penalties. The reduction of message passing complexity from quadratic to linear, described in Section \ref{sec:min-convolutions}, applies to such functions as well~\cite{ChenKoltun2014}. The highly efficient min-convolution algorithm described in Section \ref{sec:SMAWK} will assume that the function $\rho$ is convex. Other linear-time algorithms can be used if this assumption doesn't hold~\cite{ChenKoltun2014}.

Note that the regularization term (\ref{eq:reg-form}) couples the horizontal and vertical components of the flow. We apply a Laplace weight to attenuate the regularization along color discontinuities:
$$
w_{p,q} = \exp\left(-\frac{\|I_1(p)-I_2(q)\|}{\beta}\right).
$$




\section{Optimization}
\label{sec:optimization}


Objective (\ref{eqn:HS}) is a discrete Markov random field with a two-dimensional label space \cite{Blake2011}. At first glance, global optimization of this model may appear intractable. The number of nodes and the number of labels are both in the tens of thousands. The most sophisticated prior application of discrete optimization to this problem resorted to pruning of the label space to bring the size of the problem under control \cite{Menze2015}. We show that the full problem is tractable.

\subsection{Message passing algorithm}
\label{sec:TRW-S}

The label space of the model is $[-\varsigma,\varsigma]^2\cap\Ze^2$. Let \mbox{$N = |\Omega|$} be the number of nodes and let $M = (2\varsigma +1)^2$ be the size of the label space.
To optimize the model, we use TRW-S, which optimizes the dual of a natural linear programming relaxation of the problem \cite{Wainwright2005,Kolmogorov2006}. We choose TRW-S due to its effectiveness in optimizing models with large label spaces \cite{Szeliski2008,ChenKoltun2015}. Note that TRW-S optimizes the dual objective and will generally not yield the optimal solution to the primal problem.

For notational simplicity, we omit the weights $w_{p,q}$, although incorporating the weights into the presented method is straightforward. We first write down objective (\ref{eqn:HS}) as an equivalent discrete labeling problem. Let $L=[-\varsigma,\varsigma]^2\cap\Ze^2$ be the candidate flow vectors and \mbox{$|L|=M=(2\varsigma+1)^2$}. We optimize the following objective with respect to the labeling $\ll:\Omega\rightarrow L$:
\begin{eqnarray}
\btheta(\ll)&=&\sum_{p \in \Omega}{\theta_p(l_p)}+\sum_{\{p,q\}\in \N}{\theta_{pq}(l_p,l_q)}, \label{eqn:MRF}\\
\theta_p(s)&=&\rho_D(p,s,I_1,I_2),\nonumber\\
\theta_{pq}(s,t)&=&\rho_S\left(s-t\right).\nonumber
\end{eqnarray}
$\btheta$ are the potentials for the data and pairwise terms. Instead of directly minimizing objective (\ref{eqn:MRF}), TRW-S maximizes a lower bound that arises from a reparametrization of $\btheta$. $\tilde\btheta$ is said to be a reparametrization of $\btheta$ iff \mbox{$\tilde\btheta(\ll)\equiv\btheta(\ll)$}. TRW-S constructs reparameterizations of the following form:
\begin{eqnarray*}
\tilde\theta_p(s)&=&\theta_p(s)+\sum_{\{p,q\}\in \N}{m_{q\rightarrow p}(s)}\\
\tilde\theta_{pq}(s,t)&=&\theta_{pq}(s,t)-m_{q\rightarrow p}(s)-m_{p\rightarrow q}(t),
\end{eqnarray*}
where $m_{p\rightarrow q}$ is a message (a vector of size $M$) that pixel $p$ sends to its neighbor $q$ \cite{Kolmogorov2006}. Given a reparameterization $\tilde\btheta$, a lower bound for $\btheta(\ll)$ can be obtained by summing the minima of all the potentials:
\begin{equation*}
\Phi(\tilde\btheta)=\sum_{p\in \Omega}{\min_{s}{\tilde\theta_p(s)}}+\sum_{\{p,q\}\in \N}{\min_{s,t}\tilde\theta_{pq}(s,t)}.
\end{equation*}
TRW-S maximizes $\Phi(\tilde\btheta)$ by first iterating over all pixels in a given order (e.g., in scanline order or by sweeping a diagonal wavefront from one corner of the image to its antipode). When pixel $p$ is reached, the following message update rule is applied to each $m_{p\rightarrow q}$ for which $q$ has not been visited yet:
\begin{eqnarray}
m_{p\rightarrow q}(t)=&\min\limits_{s}&\biggl(\frac{1}{2}\Bigl(\theta_p(s)+\sum_{r}{m_{r\rightarrow p}(s)}\Bigr) \nonumber\\
 && -m_{q\rightarrow p}(s)+\theta_{pq}(s,t)\biggr).
\label{eqn:message_udpate}
\end{eqnarray}
The pixels are then visited in reverse order with analogous message update rules. This completes one forward-backward pass, considered to be one iteration. A number of iterations are performed.

Given updated messages $\mathbf{m}$, we can compute a solution $\ll$ greedily \cite{Kolmogorov2006}. We determine the labels sequentially in a given order. Upon reaching pixel $p$, we choose a label assignment $l_p$ that minimizes $\theta_p(l_p)+\sum_{q<p}{\theta_{pq}(l_p,l_q)}+\sum_{p<q}{m_{q\rightarrow p}(l_p)}$, where $p<q$ means that $p$ precedes $q$ in the order.

\subsection{Complexity reduction}
\label{sec:min-convolutions}

A brute-force implementation of a message update requires $O(M^2)$ operations as there are $M$ elements in each message and computing each element directly requires $O(M)$ operations according to update rule (\ref{eqn:message_udpate}). We now show that a message update can be performed using $O(M)$ operations in our model. This builds on the min-convolution acceleration scheme developed by Felzenszwalb and Huttenlocher \cite{FelzenszwalbHuttenlocher2006,FelzenszwalbHuttenlocher2012}. A general treatment of the one-dimensional case was presented by Chen and Koltun \cite{ChenKoltun2014}.

We begin by rewriting the message update rule (\ref{eqn:message_udpate}) as
\begin{eqnarray}
m_{p\rightarrow q}(t)&=&\min_{s} \big(\phi_{pq}(s)+\theta_{pq}(s,t)\big),\label{eqn:update_rule}\\
\phi_{pq}(s)&=&\frac{1}{2}\Bigl(\theta_p(s)+\sum_{r}{m_{r\rightarrow p}(s)}\Bigr)-m_{q\rightarrow p}(s).\nonumber
\end{eqnarray}
Each $\phi_{pq}(s)$ can be computed using $O(1)$ operations, thus $\phi_{pq}$ can be computed using $O(M)$ operations in total. We now show that, given $\phi_{pq}$, all elements of the message $m_{p\rightarrow q}$ can be computed in $O(M)$ operations as well. Recall that \mbox{$\theta_{pq}(s,t) = \rho_S(t-s)$} and that $\rho_S(\cdot)$ has the form given in equation (\ref{eq:reg-form}). Rearranging terms, we obtain
\begin{eqnarray*}
m_{p\rightarrow q}(t)=\min \Bigl(\mathcal{D}_{pq}(t),T_{pq}\Bigr),
\end{eqnarray*}
where
\begin{eqnarray*}
\mathcal{D}_{pq}(t)&=&\min_{s}\big(\phi_{pq}(s)+\rho(t^1-s^1)+\rho(t^2-s^2)\big),\\
T_{pq}&=&\min_s\big(\phi_{pq}(s)+\tau\big).\nonumber
\end{eqnarray*}
$T_{pq}$ can be computed using $O(M)$ operations given $\phi_{pq}$. We now show that $\mathcal{D}_{pq}$ can also be computed using $O(M)$ operations in total. Note that $s$ and $t$ are 2D vectors. Abusing notation somewhat, we can rewrite $\mathcal{D}_{pq}(t)$ as a two-dimensional min-convolution:\\
\resizebox{.93\linewidth}{!}{
  \begin{minipage}{\linewidth}
\begin{eqnarray*}
\mathcal{D}_{pq}(t^1,t^2)&=\min\limits_{s^1,s^2}& \Big(\phi_{pq}(s^1,s^2)+\rho(t^1-s^1)+\rho(t^2-s^2)\Big)\\
&=\min\limits_{s^2}&\Big(\min_{s^1}\big(\phi_{pq}(s^1,s^2)+\rho(t^1-s^1)\big)\\
&&+\ \rho(t^2-s^2)\Big).\\
\end{eqnarray*}
  \end{minipage}
}
This can be decomposed into two sets of $O(\sqrt{M})$ one-dimensional min-convolutions:
\begin{eqnarray}
\mathcal{D}_{pq}(t^1,t^2)&=&\min_{s^2}\mathcal{D}_{pq|s^2}(t^1)+\rho(t^2-s^2),\label{eqn:1d_min_conv1}\\
\mathcal{D}_{pq|s^2}(t^1)&=&\min_{s^1}\phi_{pq}(s^1,s^2)+\rho(t^1-s^1).\label{eqn:1d_min_conv2}
\end{eqnarray}
For each $s^2$, $\mathcal{D}_{pq|s^2}(t^1)$ can be computed for all $t_1$ by a 1D min-convolution. Then, for each $t^1$, $\mathcal{D}_{pq}(t^1,t^2)$ can be computed for all $t^2$ by a 1D min-convolution. Each min-convolution can be evaluated in $O(\sqrt{M})$ operations, for a total complexity of $O(M)$.

\subsection{Further acceleration}
\label{sec:SMAWK}

A min-convolution has the following general form:
\begin{equation}
h(i)=\min\limits_j{g(j)+\rho(i-j)}.
\label{eqn:min_convolution}
\end{equation}
It is well-known that the min-convolution can be computed using $O(n)$ operations, where \mbox{$n = \sqrt{M} = 2\varsigma+1$}~\cite{ChenKoltun2014}. However, commonly used algorithms require computing intersections of shifted copies of the function $\rho$. While each intersection can be computed in time $O(1)$, this computation can be numerically intensive for some penalty functions. Since this computation is in the inner loop, it can slow the optimization down. We now review an alternative algorithm that can be used to compute the min-convolution without computing intersections. This algorithm relies on the assumption that $\rho$ is convex, which is otherwise not necessary. Related algorithms are reviewed by Eppstein \cite{Eppstein1990}.


The algorithm is based on totally monotone matrix searching~\cite{Aggarwal1987}. Let $A$ be an $n\!\times\! n$ matrix, such that $A(i,j)=g(j)+\rho(i-j)$. Let $\ind_A(i)$ be the column index of the minimal element in the $i$th row of $A$. The min-convolution $h$ can be defined as $h(i)=A(i,\ind_A(i))$. The challenge is to evaluate $\ind_A$ in time $O(n)$ without explicitly constructing the matrix $A$.


The convexity of $\rho$ implies that $A$ is totally monotone. The totally monotone matrix search algorithm computes $\ind_A$ in $O(n)$ operations by divide-and-conquer. The algorithm first constructs an $\frac{n}{2}\!\times\! n$ submatrix $B$ by removing every other row of $A$. Then $B$ is reduced to an $\frac{n}{2}\!\times\! \frac{n}{2}$ submatrix $C$ by removing columns that do not contain minima of the rows of $B$. The minima $\ind_C$ are computed recursively, after which the missing elements of $\ind_A$ are filled in. As shown by Aggarwal et al.~\cite{Aggarwal1987}, all steps can be performed in time $O(n)$. Crucially, all steps can be performed without explicitly constructing $A$.

\begin{figure*}[t]
\begin{tabular}{c @{ } c @{ } c }
\includegraphics[width=0.32\textwidth]{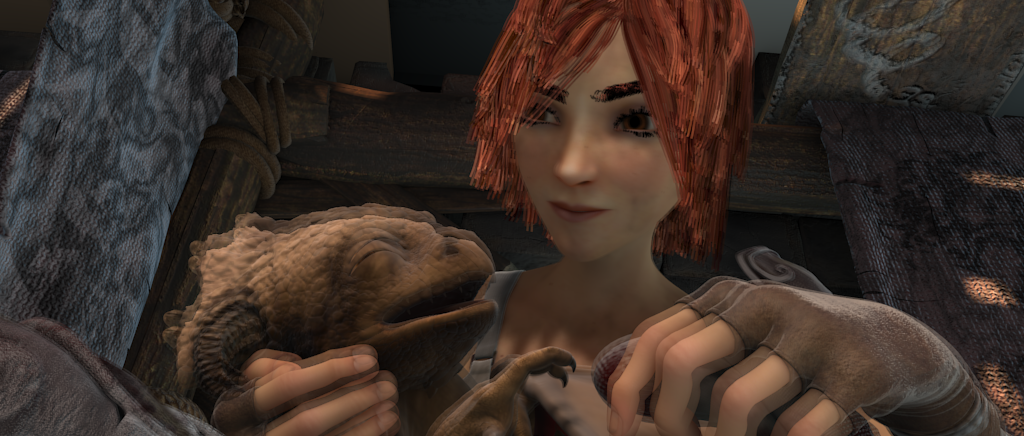}&\includegraphics[width=0.32\textwidth]{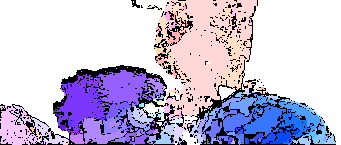}&\includegraphics[width=0.32\textwidth]{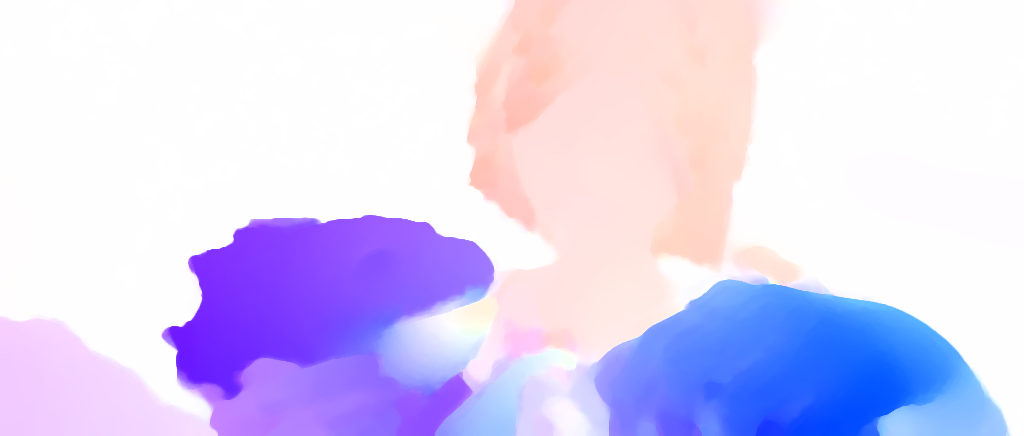}\vspace{-1mm}\\
\small{(a) Input image} & \small{(b) Before interpolation} & \small{(c) After interpolation}\vspace{1mm}\\
\includegraphics[width=0.32\textwidth]{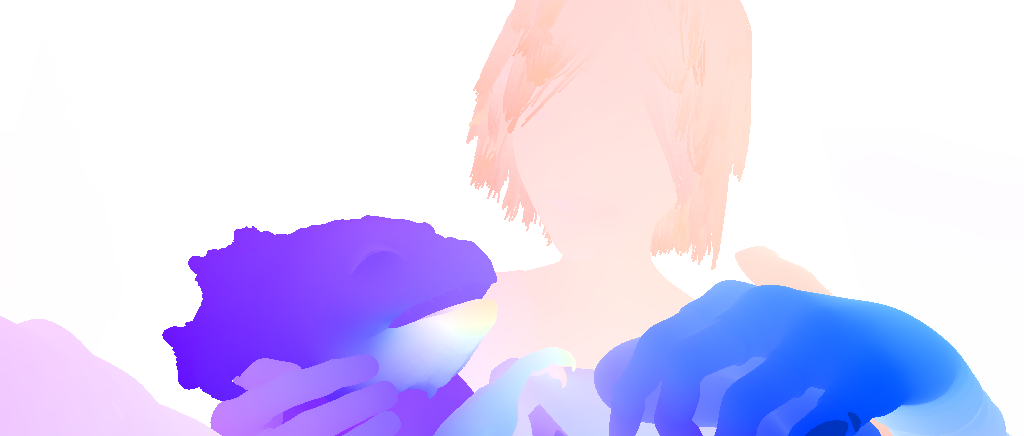}&\includegraphics[width=0.32\textwidth]{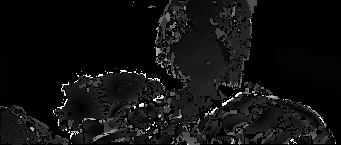}&
\includegraphics[width=0.32\textwidth]{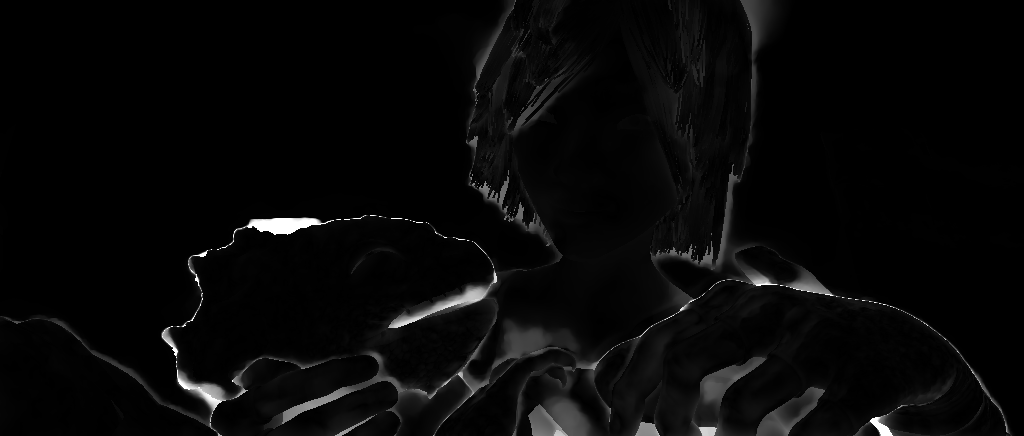}\vspace{-1mm}\\
\small{(d) Ground truth} & \small{(e) Error map for (b)} & \small{(f) Error map for (c)}\vspace{1mm}\\
\end{tabular}
\caption{Postprocessing. (a) shows the average of two input images. (b) shows the optimized flow field after forward-backward consistency checking. (c) shows the result after EpicFlow interpolation. (e) and (f) show the corresponding EPE maps, truncated at 10 pixels. }
\label{figure:postprocessing}
\vspace{-3mm}
\end{figure*}

\section{Implementation}
\label{sec:implementation}


\paragraph{Parallelization.}
To reduce wall-clock time, we implemented a parallelized TRW-S solver. This general-purpose solver along with the rest of our implementation will be made freely available. At each step of TRW-S, a pixel is ready to be processed if all of its predecessors have already been updated during the current iteration. Thus at any time there is a wavefront of pixels that can be processed in parallel. A grid can be swept diagonally. In the first step only one node can be processed, but the size of the wavefront grows rapidly and all nodes on the wavefront can be processed in parallel. This parallelization scheme has previously been explored on special-purpose hardware for stereo matching~\cite{Choi2013}. We have implemented the scheme on general-purpose processors. Our implementation is evaluated on a workstation with a 6-core Intel i7-4960X CPU. Parallelization with hyper-threading reduces the running time of each iteration of TRW-S from 256 to 39 seconds, a factor of 6.6. Since the size of the wavefront is $\Theta(\varsigma)$ for most of the iteration, increased hardware parallelism is expected to directly translate to reduction in wall-clock time. We also refer the reader to the concurrent work of Shekhovtsov et al.~\cite{Shekhovtsov2016}, who developed a parallelized energy minimization scheme that may be applicable to our setting.

\paragraph{Occlusion handling.}
Some pixels in $I_1$ may not have corresponding points in $I_2$. The computed flow field on these occlusion pixels is likely incorrect. We adopt the common tactic of forward-backward consistency checking: compute the forward flow from $I_1$ to $I_2$ and the backward flow from $I_2$ to $I_1$, and discard inconsistent matches~\cite{Ogale2005}. Given the forward flow field $\ff$ and the backward flow field $\ff'$, the following criterion is used to determine whether a match is consistent. For each pixel $p$ in $I_1$ and its match $(p+f_p)$ in $I_2$, $f_p$ is said to be consistent if there is a pair $(q+f'_q, q) \in I_1 \!\times\! I_2$ that is close to the pair $(p,p+f_p)$. Specifically, for each $f_p$, we check if there exists $f'_q$ for which
$$
\|p-(q+f'_q)\|^2+\|(p+f_p)-q\|^2<\delta.
$$
This test can be performed by finding nearest neighbors across two point sets: \mbox{$\{(p,p+f_p)\}_{p\in\Omega}$} and \mbox{$\{(q+f'_q,q)\}_{q\in\Omega}$}.

\paragraph{Postprocessing.}
We optimize the model described in Section \ref{sec:model}, remove inconsistent matches as described in the previous paragraph, and then interpolate the results to get subpixel-resolution flow. We use the EpicFlow interpolation scheme \cite{Revaud2015}, which has become a common postprocessing step in recent state-of-the-art pipelines \cite{Bailer2015,Menze2015}. Since an interpolation step is necessary to obtain subpixel-accurate flow, the discrete optimization need not operate at the highest resolution. We found that optimizing the presented model on $1/3$-resolution images still yields state-of-the-art performance. We attribute this both to the power of the presented global optimization approach and to the effectiveness of the EpicFlow interpolation scheme. The postprocessing is illustrated in Figure \ref{figure:postprocessing}.

\section{Experiments}
\label{sec:experiments}

The presented approach is implemented in Matlab, with a C++ wrapper for the parallelized TRW-S solver. Our Matlab code is less than 50 lines long, not including the general-purpose solver.

The experiments are performed on two challenging optical flow datasets, MPI Sintel \cite{Butler2012} and KITTI 2015 \cite{MenzeGeiger2015}. We use a workstation with a 6-core Intel i7-4960X 3.6GHz CPU and 64GB of RAM. The computation of the 1.3 billion values in the unary cost volume takes 51 seconds. We use $3\!\times\!3$ patches for NCC in $1/3$-resolution images. Performing 3 iterations of TRW-S using the general optimization framework described in Section \ref{sec:optimization} takes about 2 minutes on either Sintel or KITTI images, downsampled by a factor of 3, with any penalty function $\rho$. When the penalty function is the $L^1$ norm, we can accelerate the optimization further with the $L^1$ distance transform \cite{FelzenszwalbHuttenlocher2012}, which reduces the running time to about 30 seconds for 3 iterations of TRW-S. EpicFlow interpolation takes 3 seconds. For each dataset, we train the parameters on $5\%$ of the training set by grid search. We use the same parameters for the `final' and `clean' sequences in the Sintel dataset.

\begin{table*}[t]
  \centering
\begin{tabular}{|c|c|c|c|c|c|c|c|c|c|c|c|c|}
\hline
&\multicolumn{5}{|c|}{Final pass} & \multicolumn{5}{|c|}{Clean pass}\\\hline
 & all &noc &occ &d0-10 & s40+& all &noc &occ &d0-10 & s40+\\\hline
FlowFields \cite{Bailer2015}	&\bf{5.810}&\bf{2.621}&31.799&\bf{4.851}&\bf{33.890}&3.748&\bf{1.056}&25.700&2.784&23.602\\\hline
FullFlow &5.895 &2.838	&\bf{30.793} &4.905	& 35.592 & \bf{3.601} &1.296 & \bf{22.424} & 2.944 & \bf{20.612}\\\hline
DiscreteFlow \cite{Menze2015}	&6.077&2.937&31.685&5.106&36.339&\bf{3.567}&1.108&23.626&3.398&20.906\\\hline
EpicFlow \cite{Revaud2015} 		&6.285&3.060&32.564&5.205&38.021&4.115&1.360&26.595&3.660&25.859\\\hline
TF+OFM \cite{KennedyTaylor2015}	&6.727&3.388&33.929&5.544&39.761&4.917&1.874&29.735&3.676&31.391\\\hline
NNF-Local \cite{Chen2013}		&7.249&2.973&42.088&\bf{4.896}&44.866&5.386&1.397&37.896&\bf{2.722}&36.342\\\hline
PH-Flow	\cite{YangLi2015}		&7.423&3.795&36.960&5.550&44.926&4.388&1.714&26.202&3.612&27.997\\\hline
Classic+NL \cite{Sun2014}		&9.153&4.814&44.509&7.215&60.291&7.961&3.770&42.079&6.191&57.374\\\hline
\end{tabular}
\vspace{2mm}
\caption{Endpoint errors of different methods on the MPI Sintel test set. This table lists the most accurate methods and the Classic+NL baseline. `all' = over the whole image. `noc' = non-occluded pixels. `occ' = occluded pixels. `d0-10' = within 10 pixels of an occlusion boundary. \mbox{`s40+' = displacements larger than 40 pixels.}}
\label{table:Sintel}
\vspace{-3mm}
\end{table*}

\subsection{Comparison to prior work}
\label{sec:prior}

In experiments reported in this section, we use the $L^1$ norm for regularization ($\rho(x) = |x|$) and no truncation ($\tau = \infty$). This decision is motivated by the controlled experiments reported in Section \ref{sec:controlled}.

\paragraph{MPI Sintel.}
MPI Sintel is a dataset for large-displacement optical flow \cite{Butler2012}. There are two types of sequences in the dataset, clean and final. The clean sequences exhibit a variety of illumination and shading effects including specular reflectance and soft shadows. The final sequences additionally have motion blur, depth of field, and atmospheric effects.

The experimental results are provided in Table \ref{table:Sintel}. We use the $10$ metrics reported by Bailer et al.~\cite{Bailer2015}, including all, noc, occ, d0-10, and s40+ for both clean and final test sequences. All the errors are measured as endpoint error (EPE), which is the Euclidean distance between the estimated flow vector and the ground truth. Since some error metrics are extremely close for different methods, and because the average EPE is sensitive to outliers (the top methods generally have errors of $20$ to $40$ pixels on a number of challenging sequences), we highlight every method that achieves within $1\%$ of the lowest reported error as one of the top methods according to that error metric.

Our approach outperforms EpicFlow \cite{Revaud2015}, TF+OFM \cite{KennedyTaylor2015}, NNF-Local \cite{Chen2013}, PH-Flow \cite{YangLi2015}, and Classic+NL \cite{Sun2014} on almost all metrics. Our approach ranks 2nd on the key EPE-all metric for both final and clean sequences. Compared to EpicFlow, our approach reduces EPE-all by $6.2\%$ on the final sequences and by $12.5\%$ on the clean sequences.

\paragraph{KITTI 2015.}
KITTI Optical Flow 2015 is an optical flow dataset that comprises outdoor images of dynamic scenes captured from a car \cite{Geiger2013,MenzeGeiger2015}. The car is equipped with a LiDAR sensor and color cameras. Ground-truth flow is obtained by rigid alignment of the static environment and by fitting CAD models to moving objects. Ground-truth correspondences are sparse. The dataset contains $200$ training sequences and $200$ test sequences. A flow vector is considered an outlier if its endpoint error is 3 pixels or higher. Table \ref{table:KITTI_2015} lists the most accurate methods on this dataset, along with the classical Horn-Schunck algorithm for reference. Note that SOF \cite{SevillaBlack2016} was developed concurrently with our work and uses substantially more information at training time, at the cost of generality.

\begin{table}[h]
\centering
\begin{tabular}{|c|c|c|}
\hline
 & all & non-occluded \\\hline
SOF \cite{SevillaBlack2016} & \bf{16.81\%} & \bf{10.86\%} \\ \hline
DiscreteFlow \cite{Menze2015} & 22.38\% & 12.18\%  \\ \hline
FullFlow	                  & 24.26\% & 15.35\%  \\ \hline
EpicFlow \cite{Revaud2015}	  & $27.10\%$ & $17.61\%$  \\ \hline
DeepFlow \cite{Weinzaepfel2013}  & $29.18\%$ & $19.15\%$  \\ \hline
Horn-Schunck \cite{Sun2014}	  & $42.18\%$ & $34.13\%$  \\ \hline
\end{tabular}
\vspace{2mm}
\caption{Accuracy of different methods on the KITTI 2015 test set. This table lists the percentage of outliers on all pixels and on non-occluded pixels.}
\label{table:KITTI_2015}
\vspace{-3mm}
\end{table}

\begin{figure*}[t]
\centering
\begin{tabular}{c@{ }c@{\hspace{0.5mm}}c@{\hspace{0.5mm}}c@{\hspace{0.5mm}}c}

\rotatebox{90}{\hspace{-11mm}\footnotesize{Sintel scene 1}}&\includegraphics[width=0.24\textwidth]{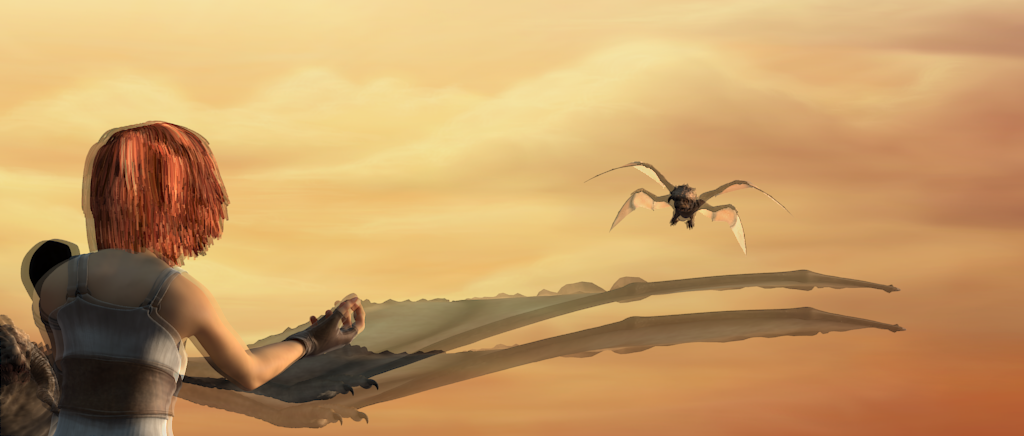}&\includegraphics[width=0.24\textwidth]{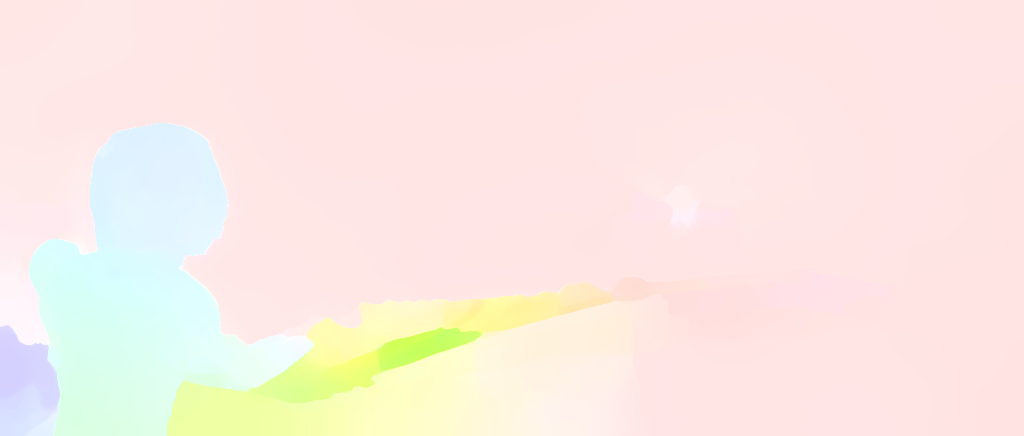}&\includegraphics[width=0.24\textwidth]{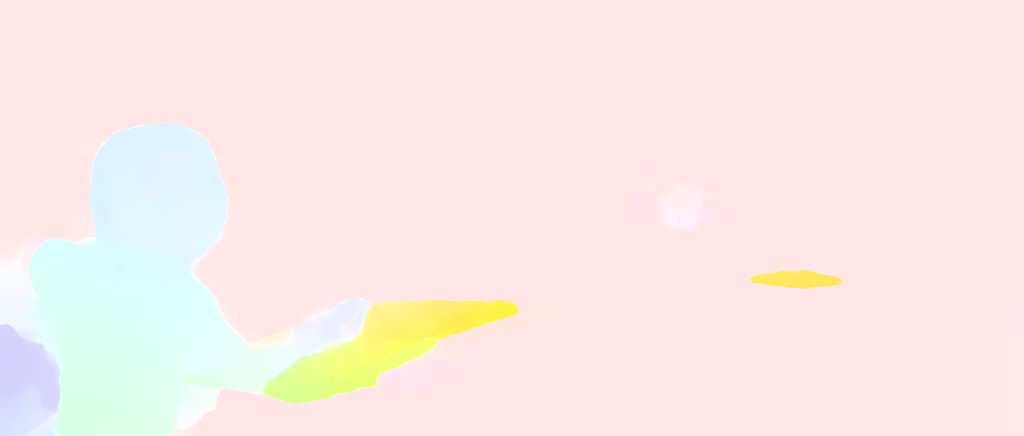}&\includegraphics[width=0.24\textwidth]{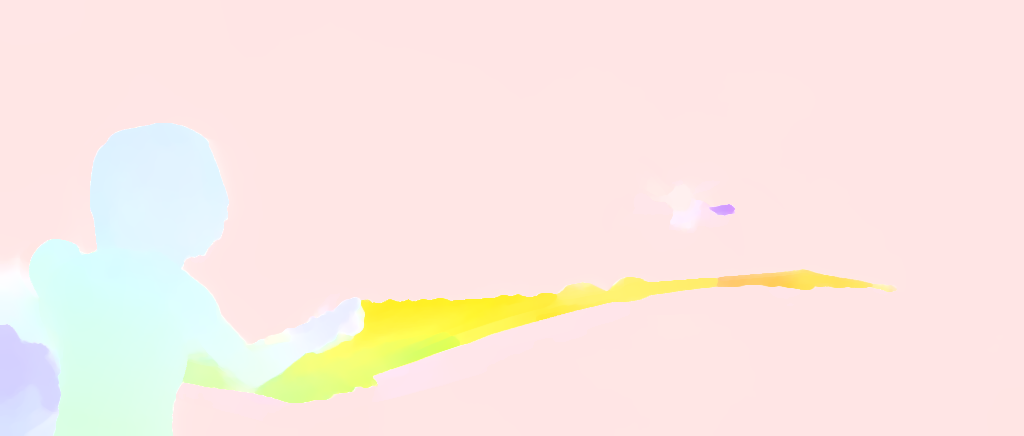}\\
&\small{(a) Input image} & \small{(b) EpicFlow \cite{Revaud2015}} & \small{(c) DiscreteFlow \cite{Menze2015}} & \small{(d) FullFlow} \\
&\includegraphics[width=0.24\textwidth]{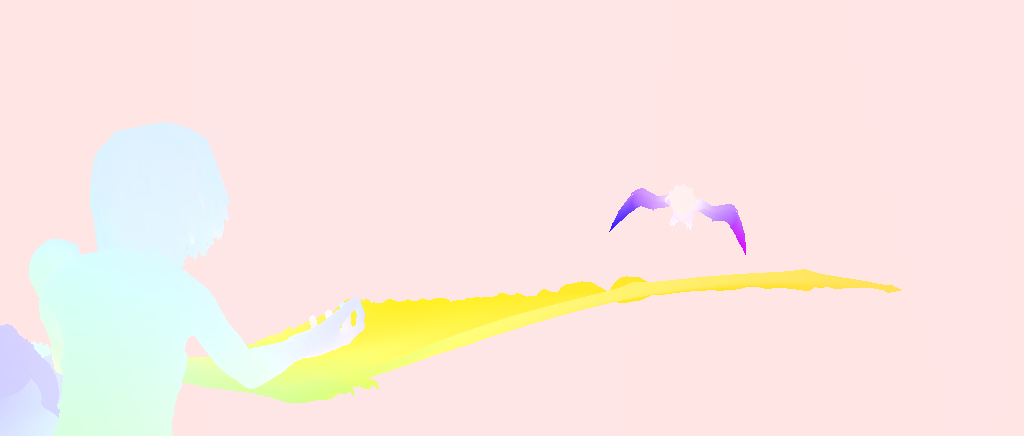}&\includegraphics[width=0.24\textwidth]{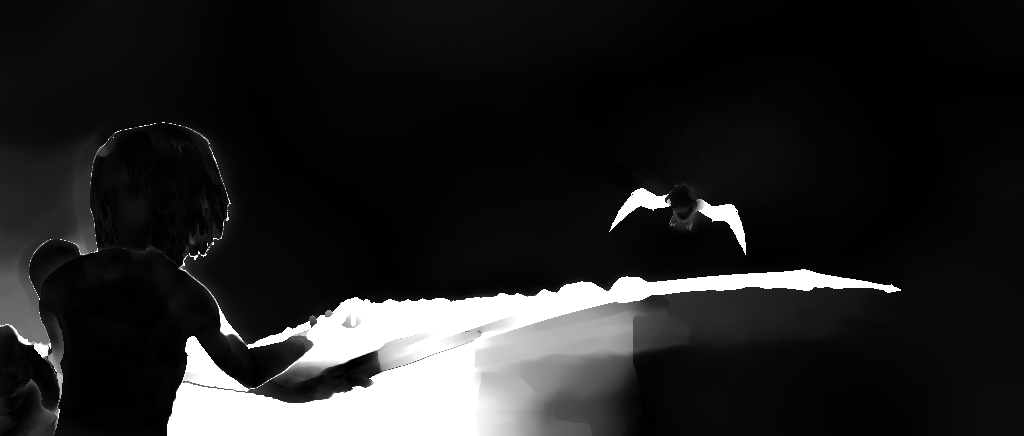}&\includegraphics[width=0.24\textwidth]{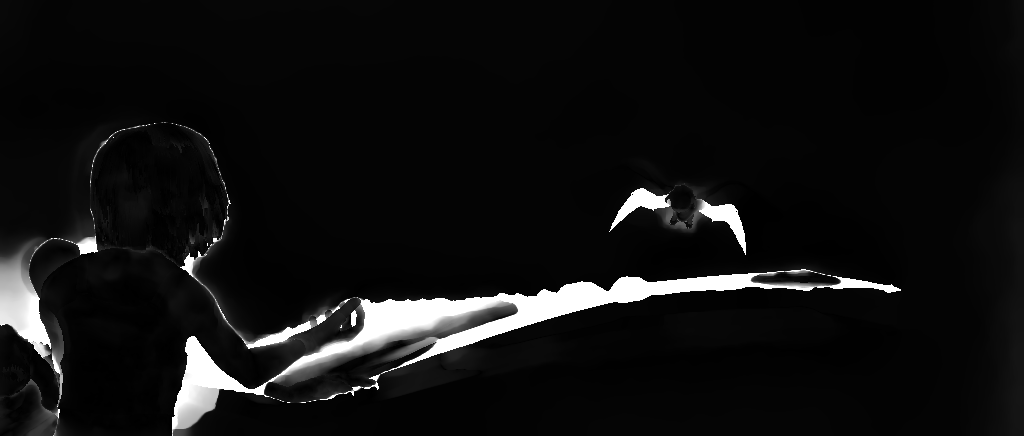}&\includegraphics[width=0.24\textwidth]{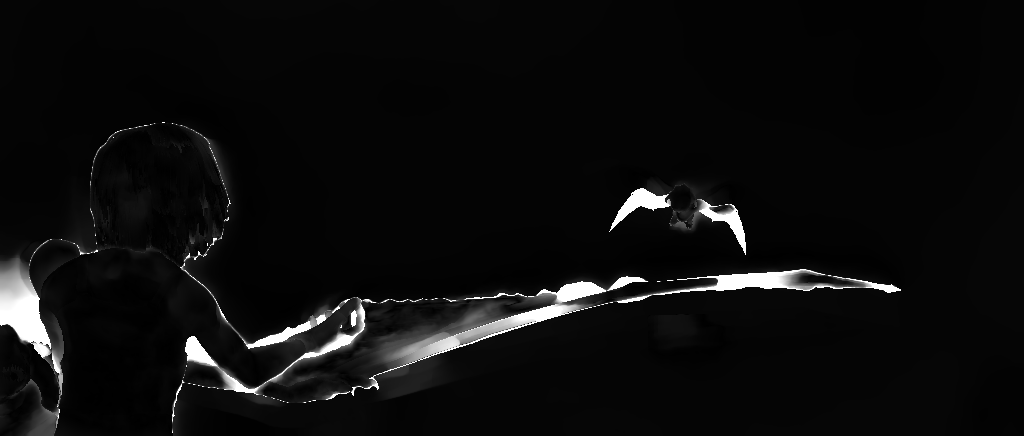}\\
&\small{(e) Ground truth} & \small{(f) Error map for (b)} & \small{(g) Error map for (c) } & \small{(h) Error map for (d)} \vspace{0.2mm}\\

\rotatebox{90}{\hspace{-7mm}\footnotesize{Sintel scene 2}}&\includegraphics[width=0.24\textwidth]{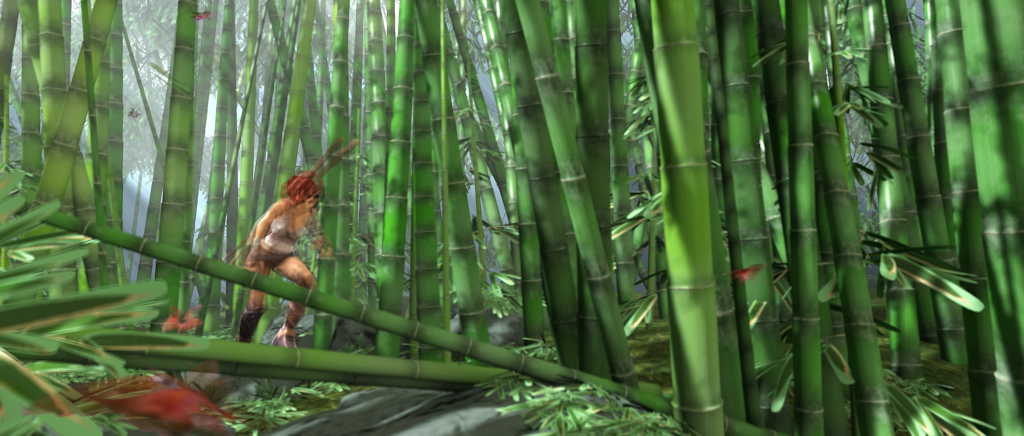}&\includegraphics[width=0.24\textwidth]{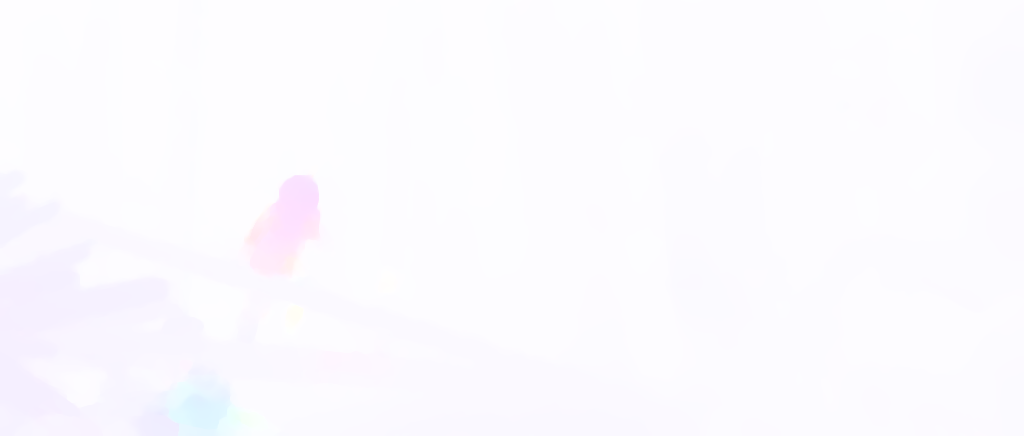}&\includegraphics[width=0.24\textwidth]{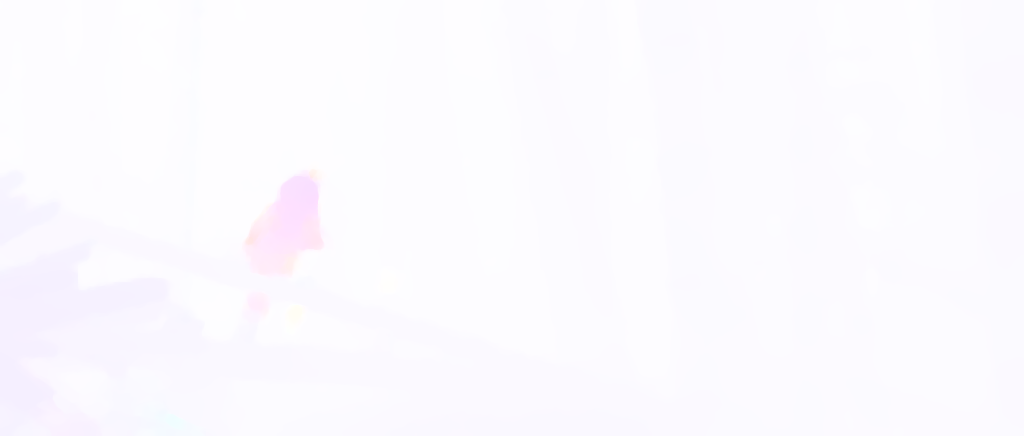}&\includegraphics[width=0.24\textwidth]{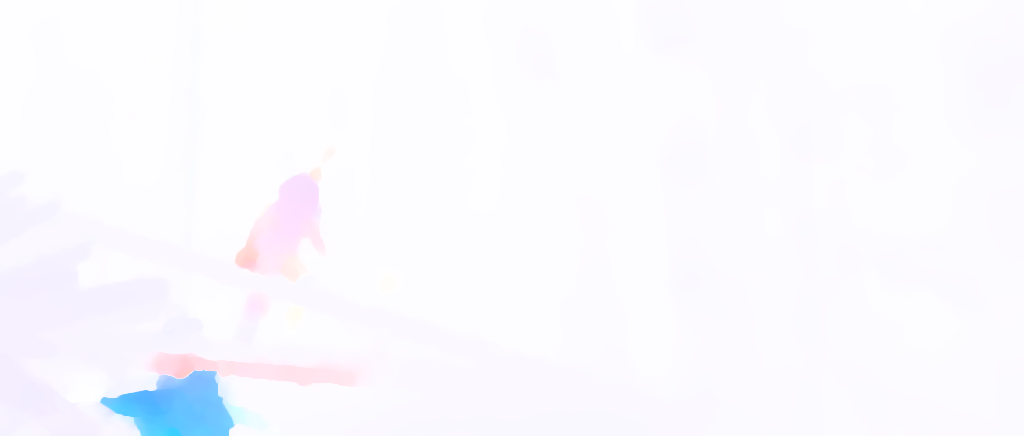}\vspace{-0.8mm}\\
&\includegraphics[width=0.24\textwidth]{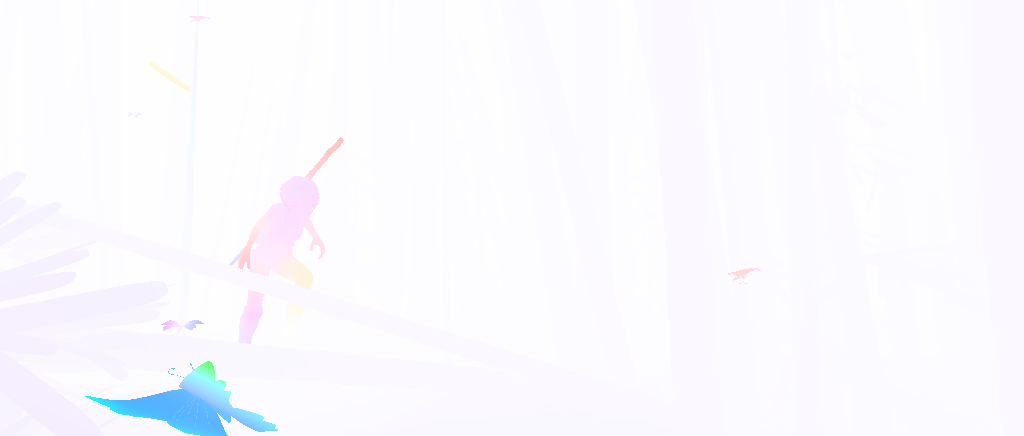}&\includegraphics[width=0.24\textwidth]{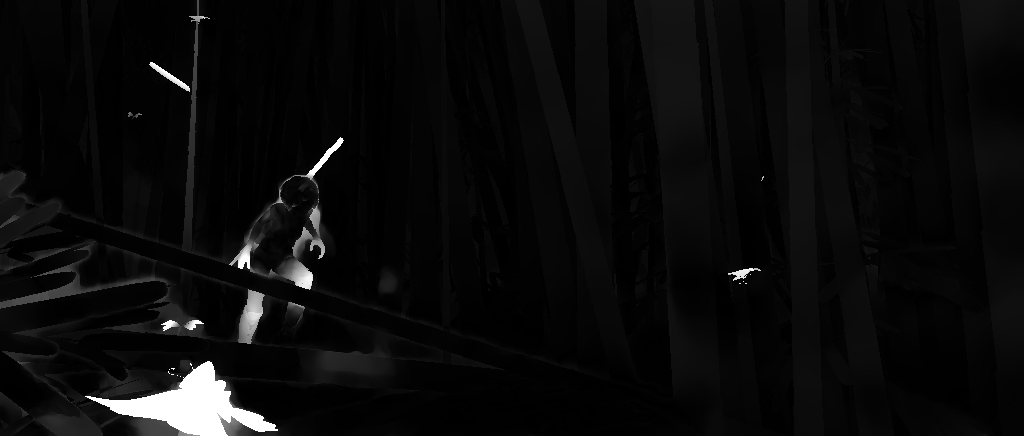}&\includegraphics[width=0.24\textwidth]{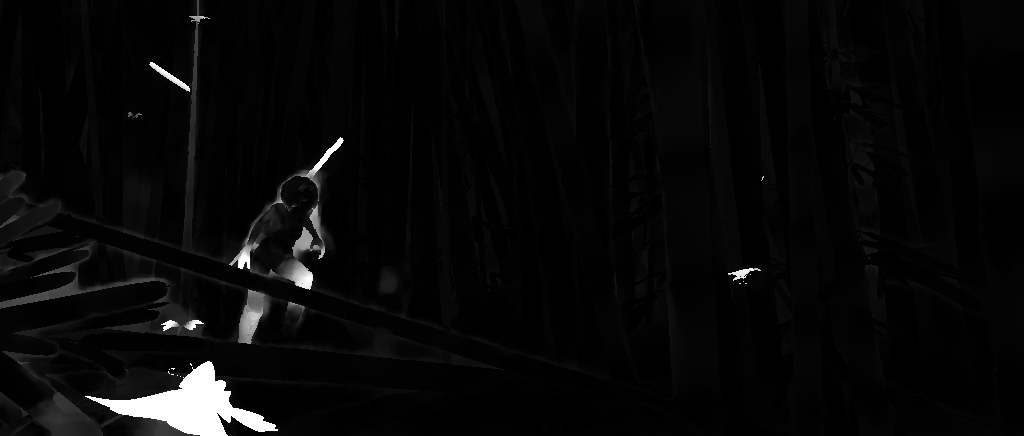}&\includegraphics[width=0.24\textwidth]{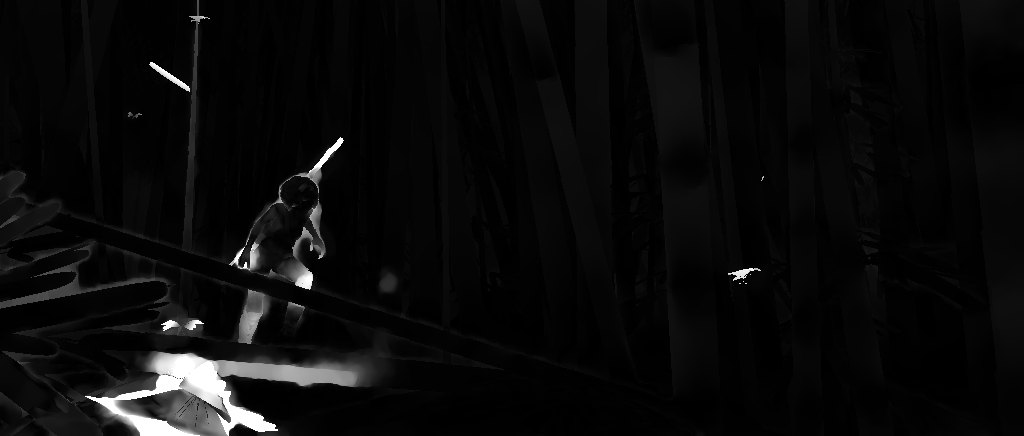}\vspace{1.2mm}\\

\rotatebox{90}{\hspace{-7mm}\footnotesize{Sintel scene 3}}&\includegraphics[width=0.24\textwidth]{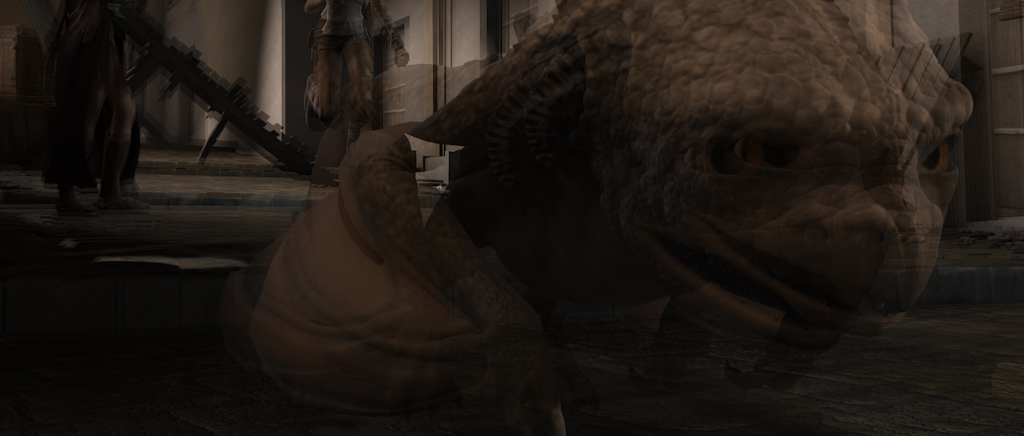}&\includegraphics[width=0.24\textwidth]{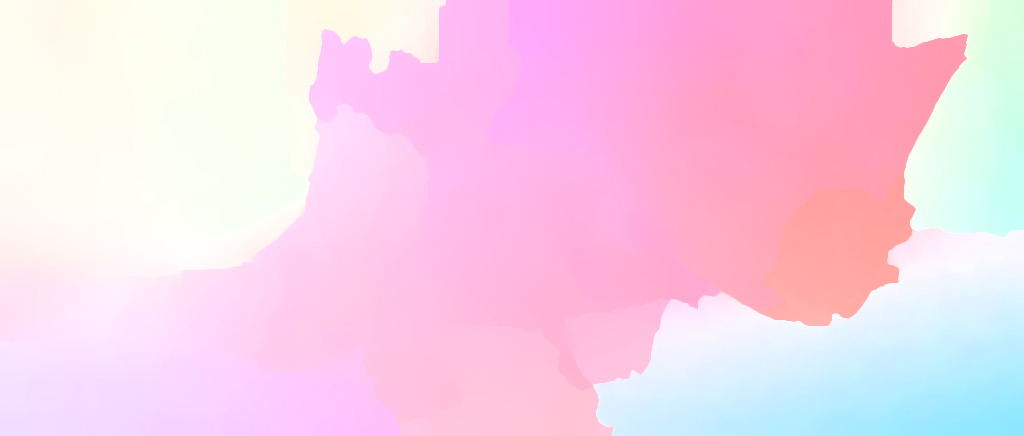}&\includegraphics[width=0.24\textwidth]{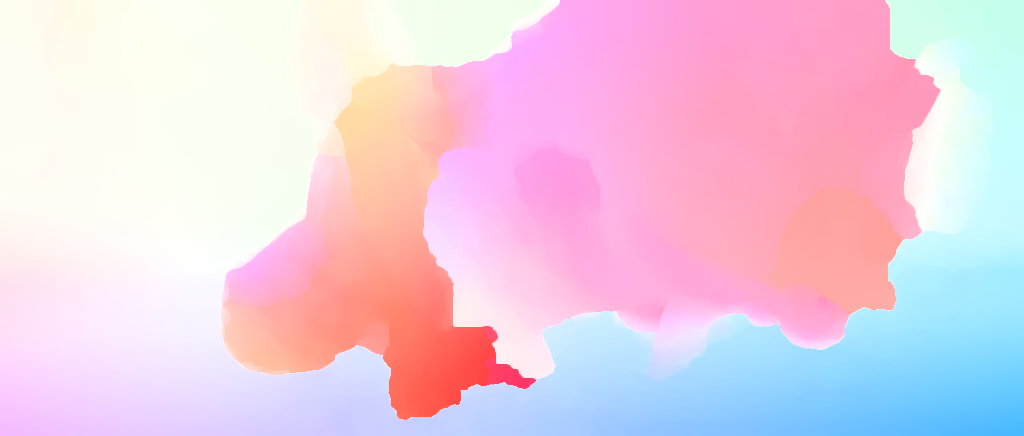}&\includegraphics[width=0.24\textwidth]{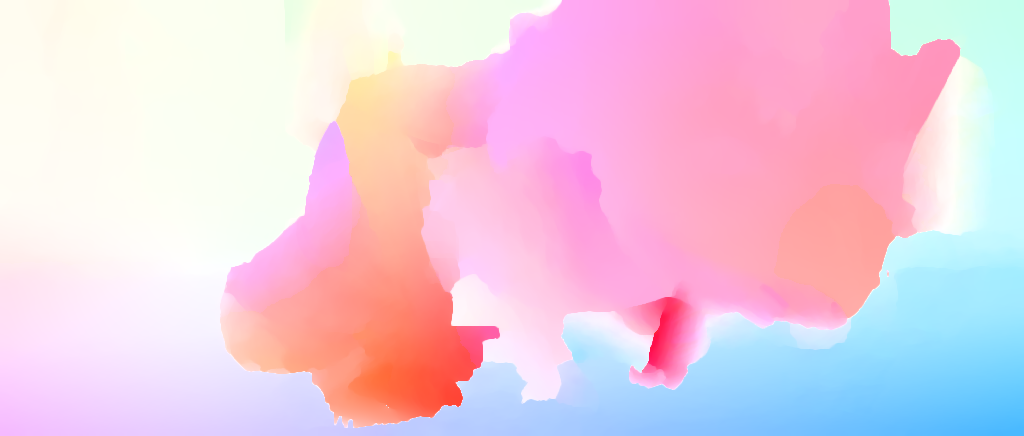}\vspace{-0.8mm}\\
&\includegraphics[width=0.24\textwidth]{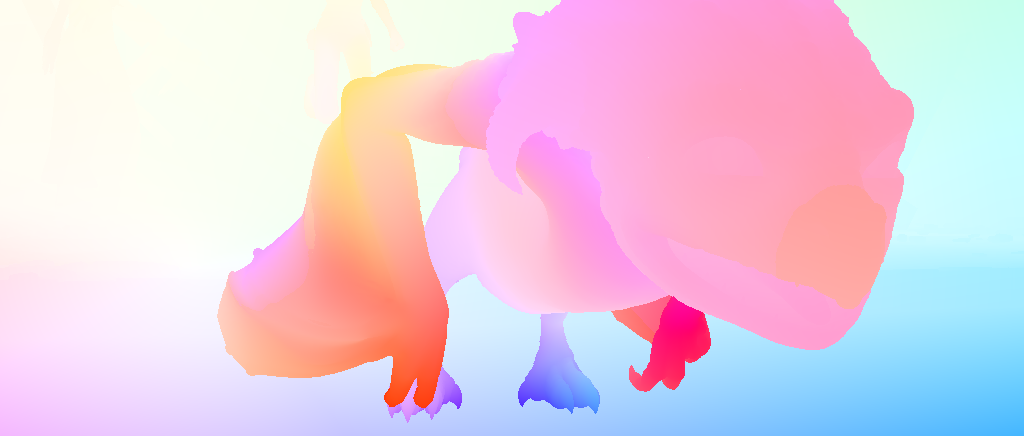}&\includegraphics[width=0.24\textwidth]{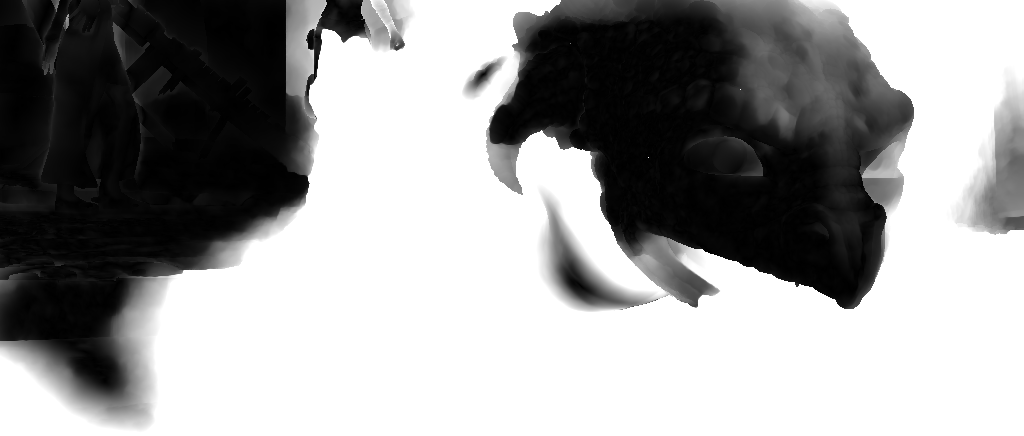}&\includegraphics[width=0.24\textwidth]{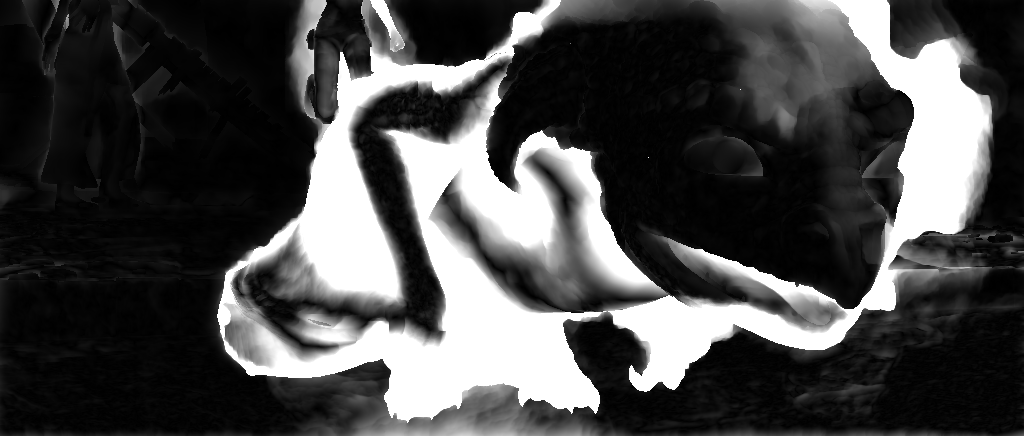}&\includegraphics[width=0.24\textwidth]{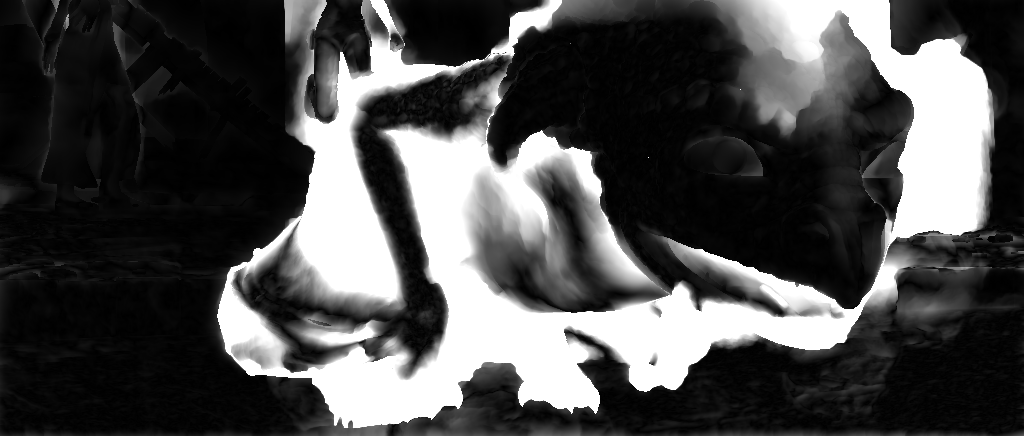}\vspace{1.2mm}\\

\rotatebox{90}{\hspace{-8mm}\footnotesize{KITTI scene 1}}&\includegraphics[width=0.24\textwidth]{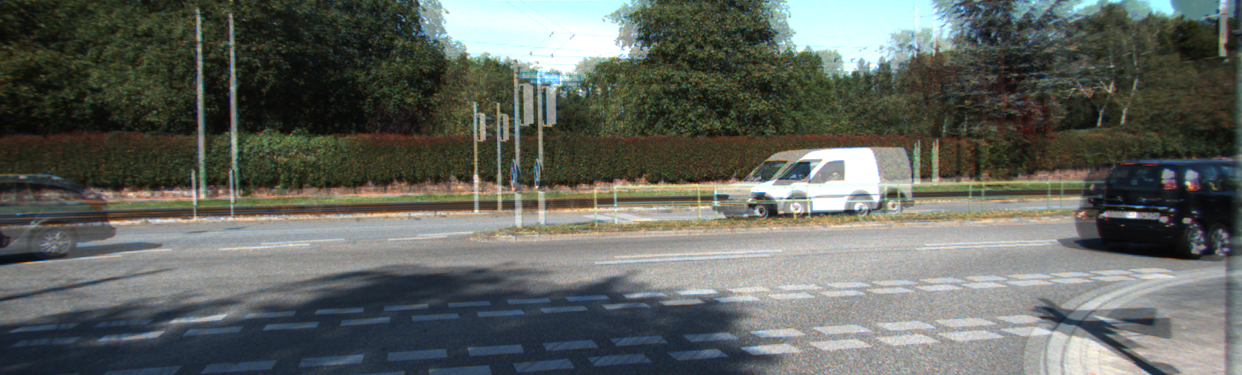}&\includegraphics[width=0.24\textwidth]{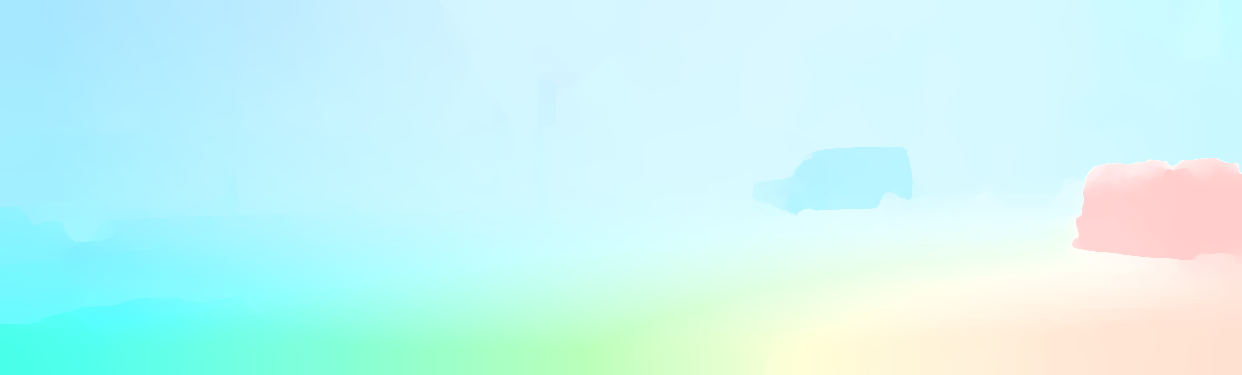}&\includegraphics[width=0.24\textwidth]{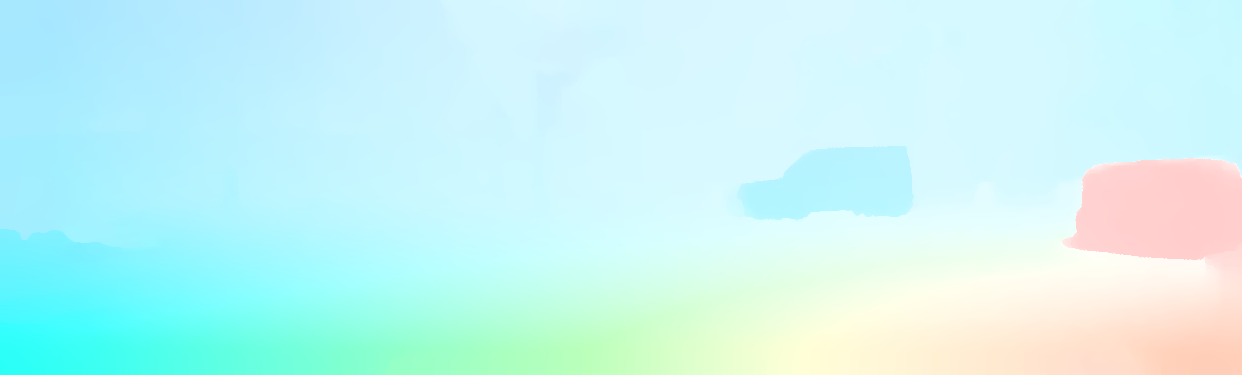}&\includegraphics[width=0.24\textwidth]{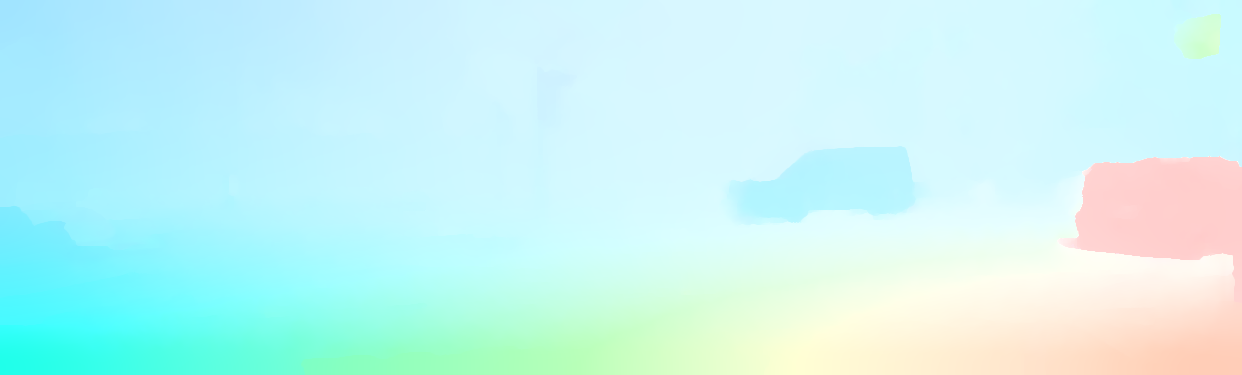}\vspace{-0.8mm}\\
&\includegraphics[width=0.24\textwidth]{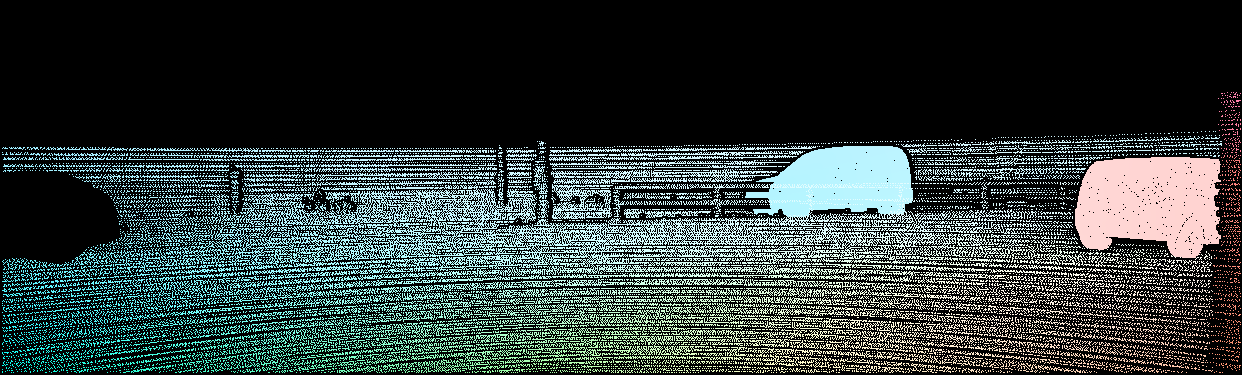}&\includegraphics[width=0.24\textwidth]{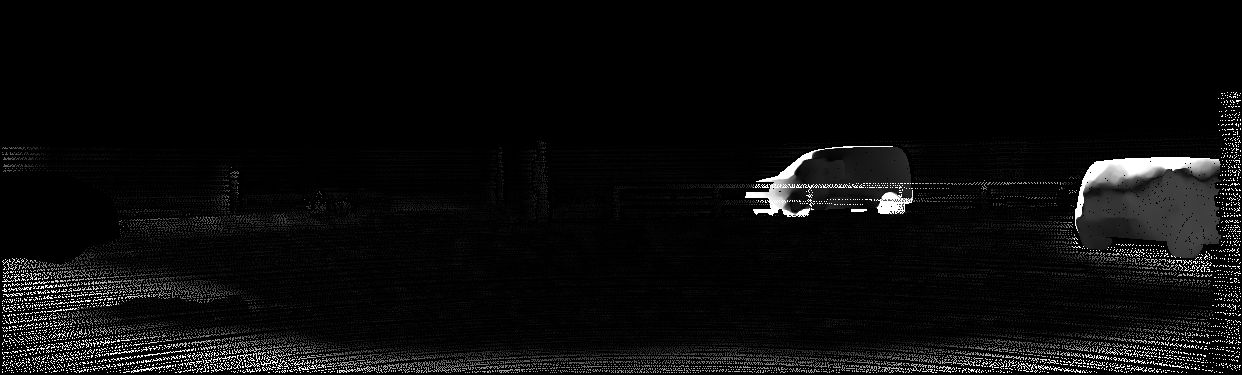}&\includegraphics[width=0.24\textwidth]{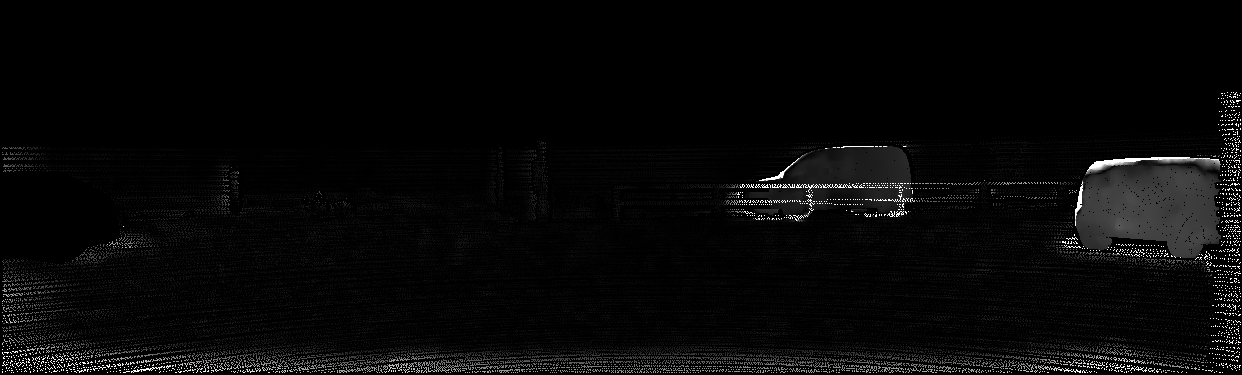}&\includegraphics[width=0.24\textwidth]{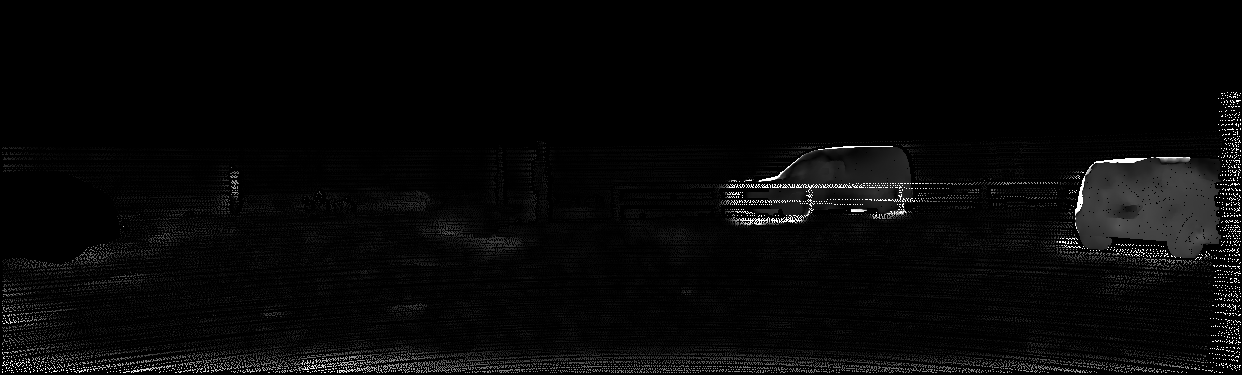}\vspace{1.2mm}\\

\rotatebox{90}{\hspace{-8mm}\footnotesize{KITTI scene 2}}&\includegraphics[width=0.24\textwidth]{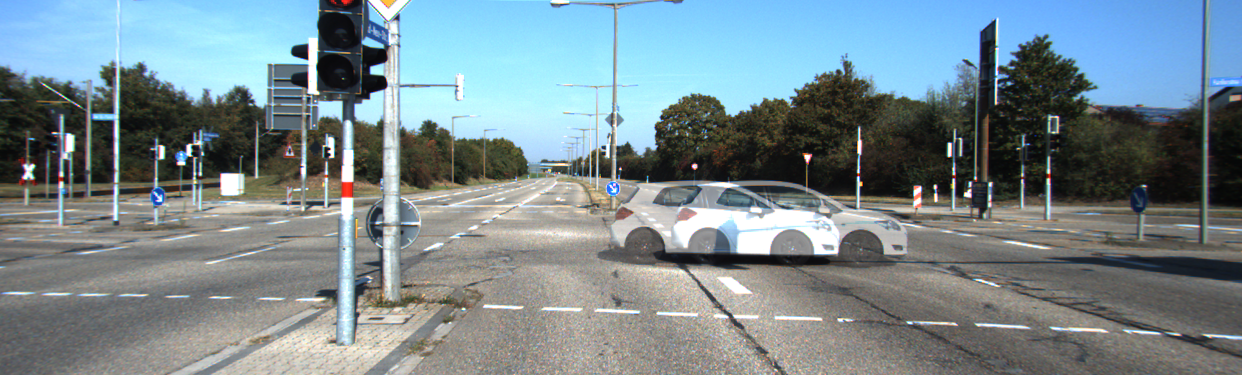}&\includegraphics[width=0.24\textwidth]{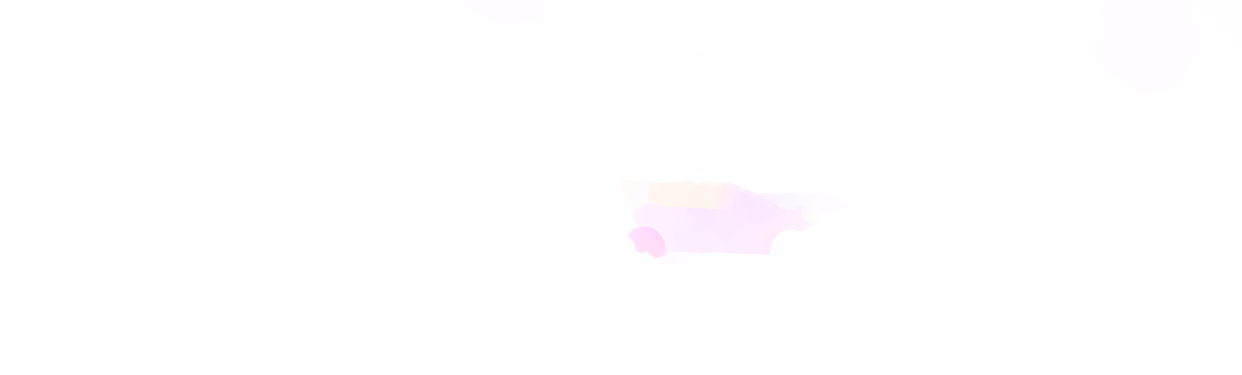}&\includegraphics[width=0.24\textwidth]{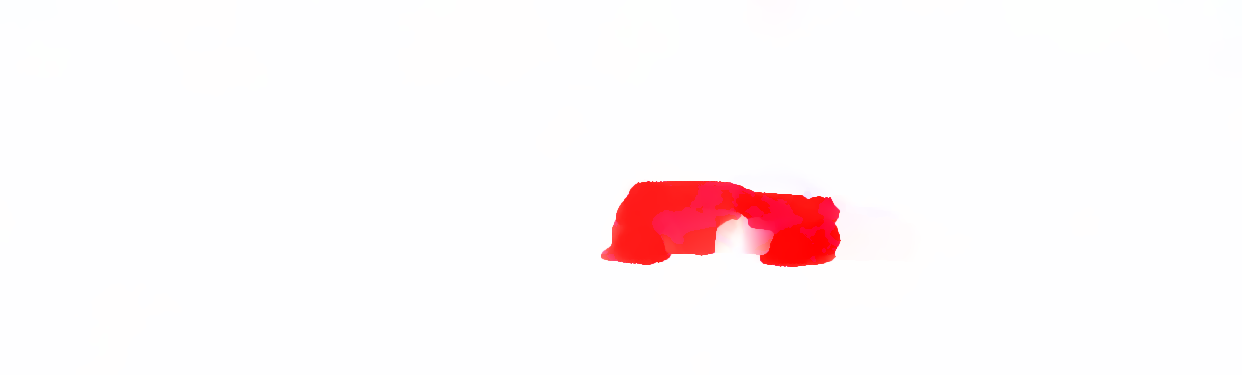}&\includegraphics[width=0.24\textwidth]{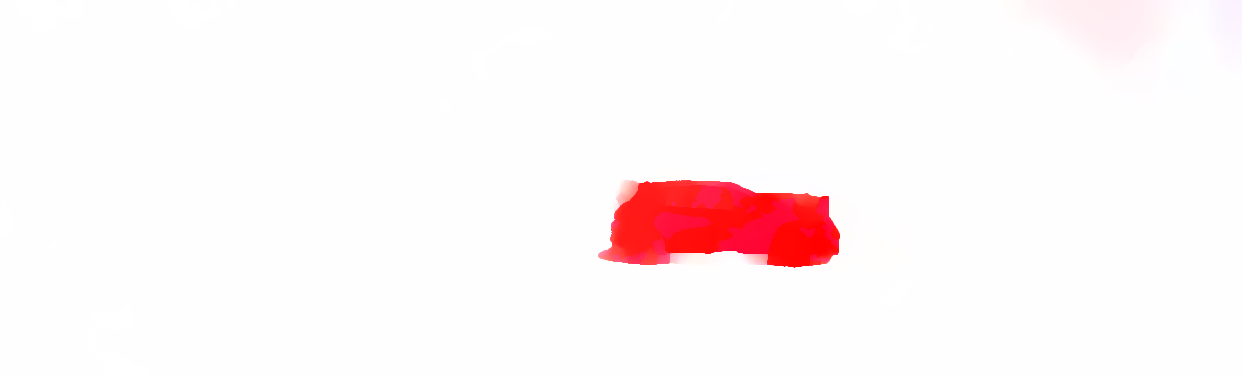}\vspace{-0.8mm}\\
&\includegraphics[width=0.24\textwidth]{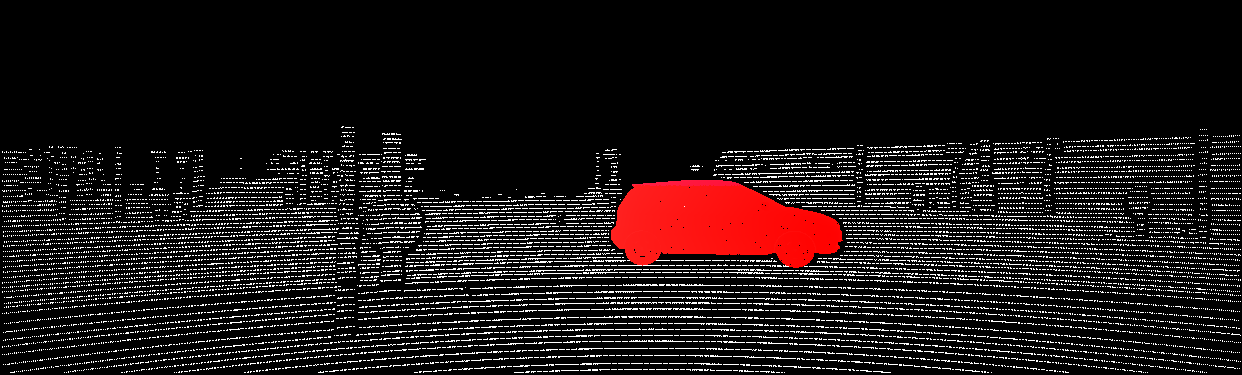}&\includegraphics[width=0.24\textwidth]{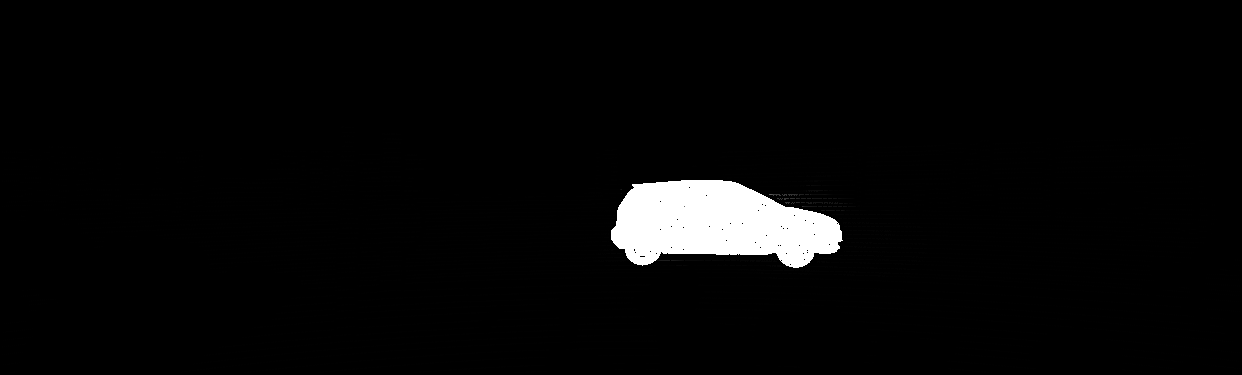}&\includegraphics[width=0.24\textwidth]{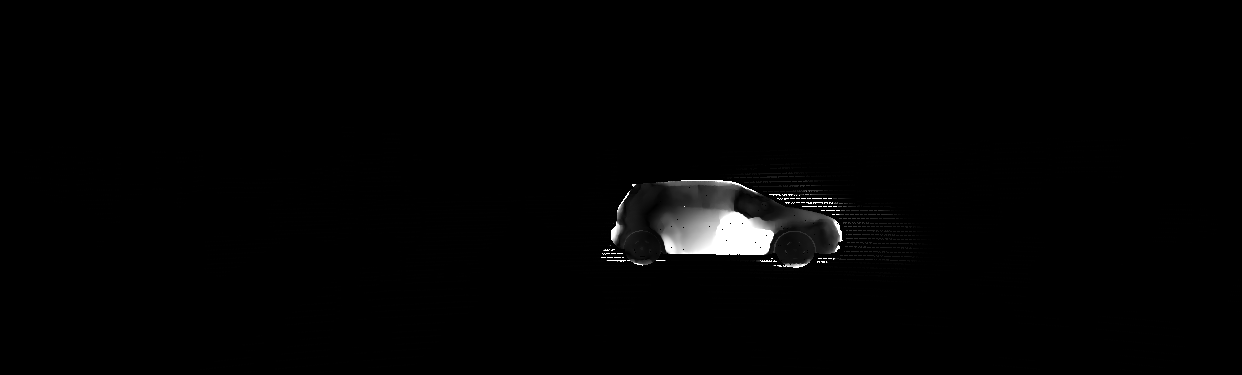}&\includegraphics[width=0.24\textwidth]{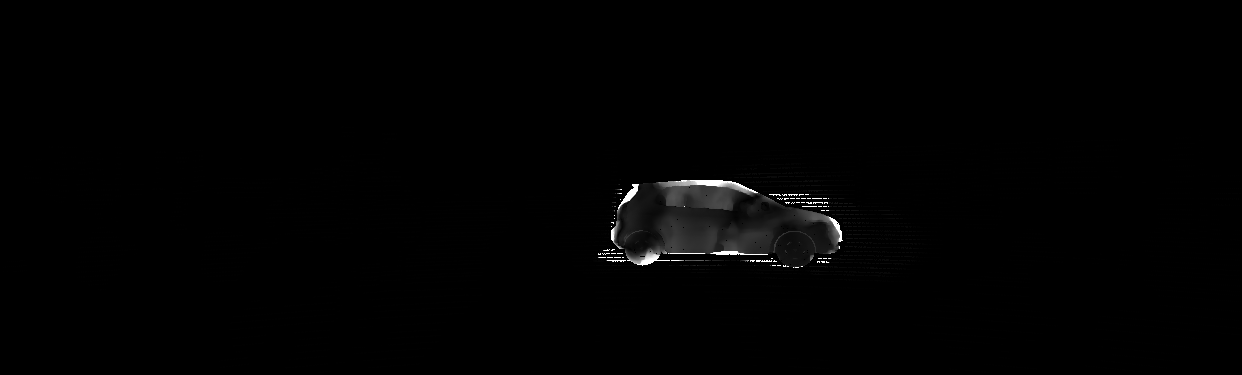}\vspace{1.2mm}\\

\rotatebox{90}{\hspace{-8mm}\footnotesize{KITTI scene 3}}&\includegraphics[width=0.24\textwidth]{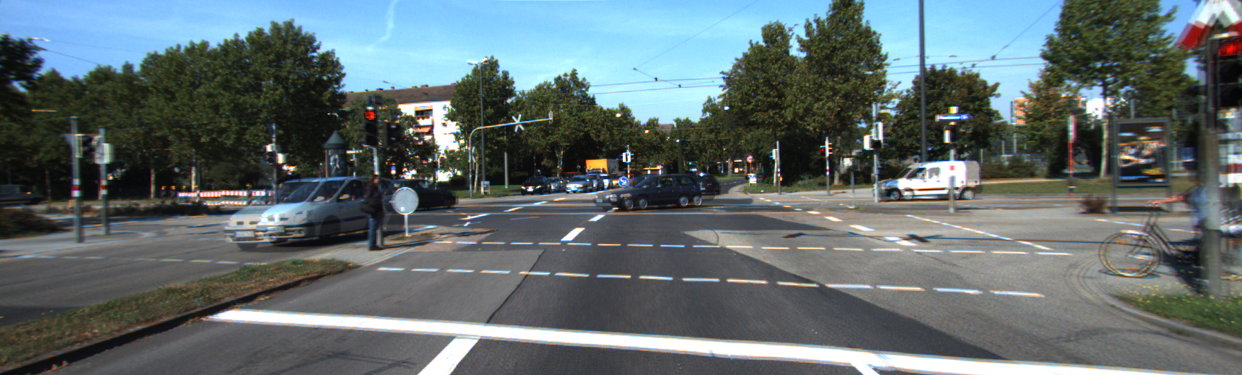}&\includegraphics[width=0.24\textwidth]{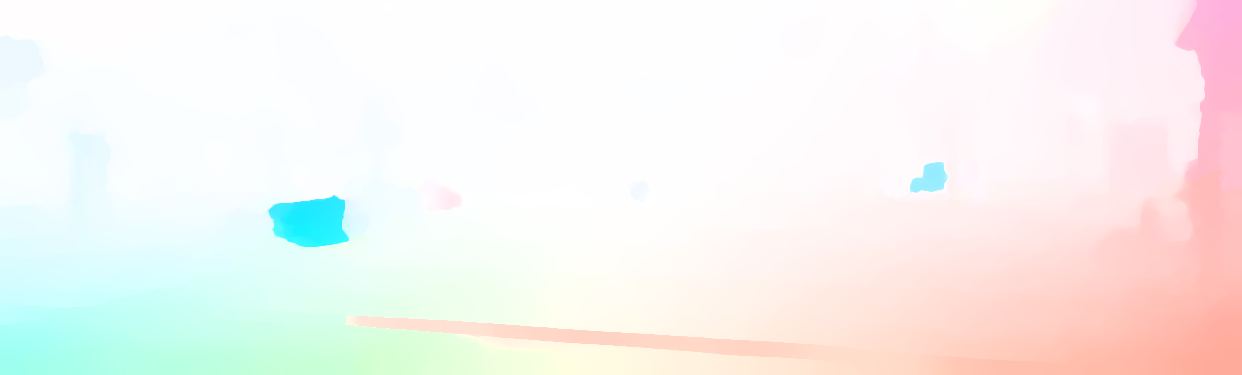}&\includegraphics[width=0.24\textwidth]{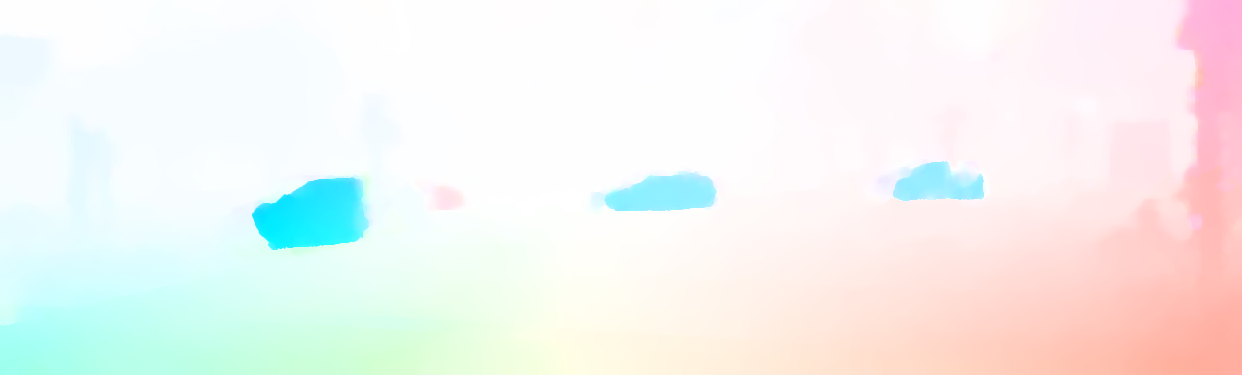}&\includegraphics[width=0.24\textwidth]{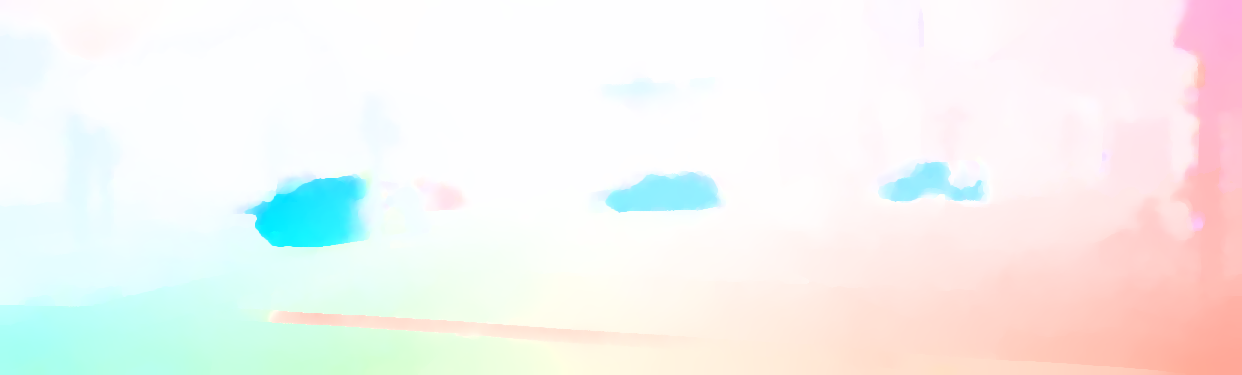}\vspace{-0.8mm}\\
&\includegraphics[width=0.24\textwidth]{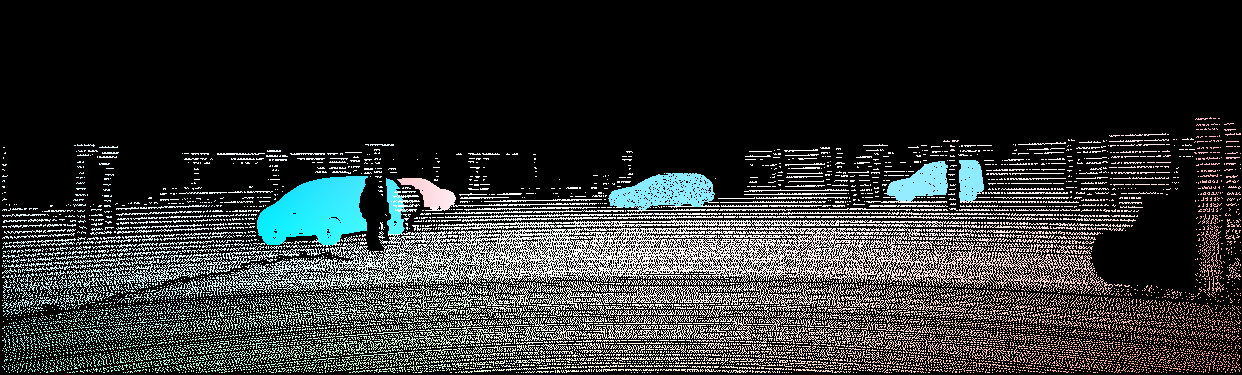}&\includegraphics[width=0.24\textwidth]{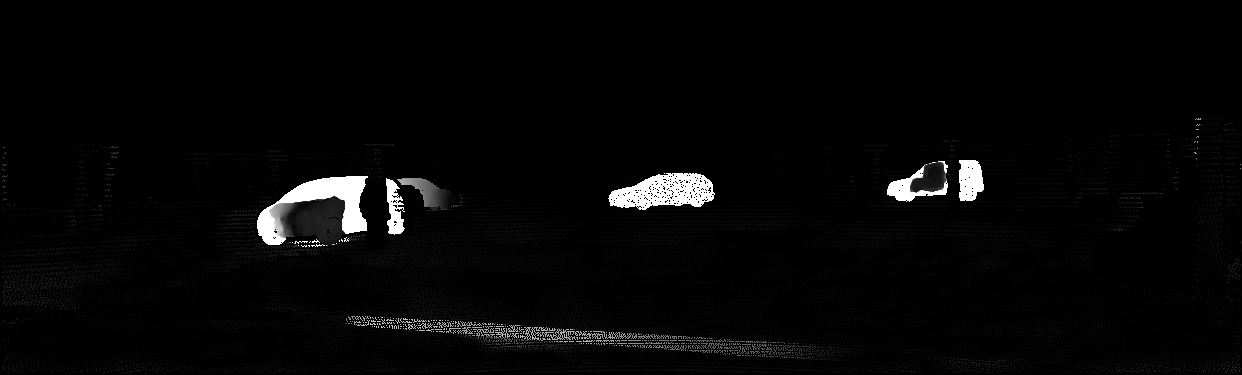}&\includegraphics[width=0.24\textwidth]{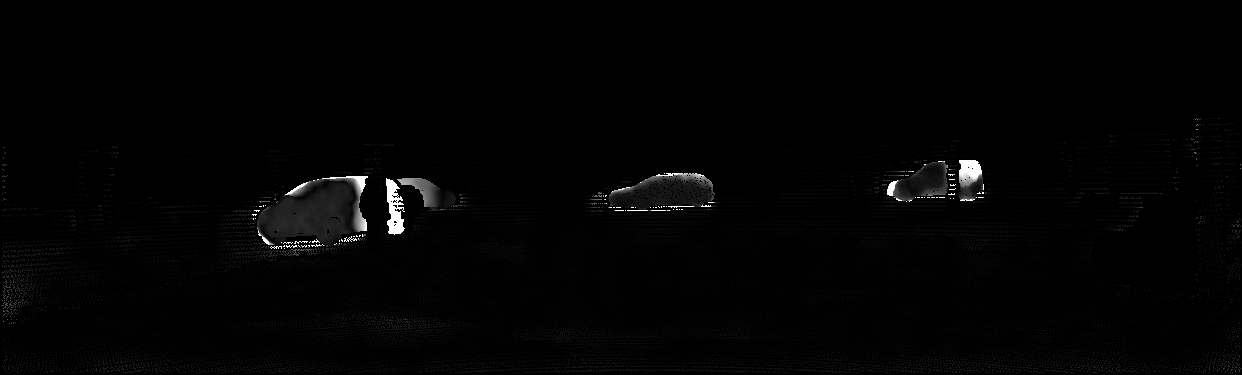}&\includegraphics[width=0.24\textwidth]{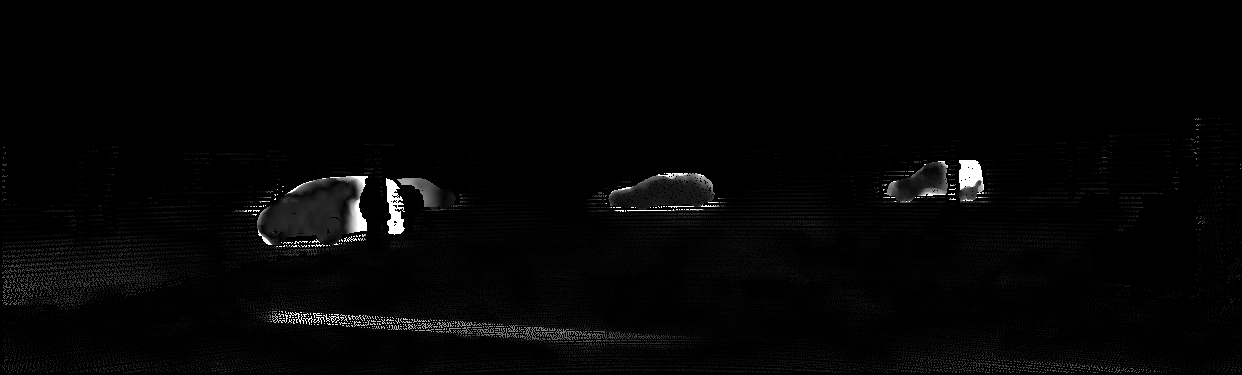}\\

\end{tabular}
\caption{Qualitative comparison on three scenes from the MPI Sintel training set (top) and three scenes from the KITTI 2015 training set (bottom). Two rows of images per scene. (a) shows the average of two input images. (b-d) show the flow fields, and (f-h) show the corresponding EPE maps, truncated at 10 pixels. The arrangement of images is the same for each scene.}
\label{figure:qualitative}
\end{figure*}

\paragraph{Qualitative results.}
In Figure \ref{figure:qualitative}, we compare our visual results to EpicFlow and DiscreteFlow on three scenes from MPI Sintel and three scenes from KITTI 2015. On MPI Sintel, our approach performs well on regions with large displacements (we rank first on s40+ in Table \ref{table:Sintel}). This is also reflected in the visual results. See the flapping wings in scene 1 and the flying butterfly in scene 2.
In scene 3, all three methods fail but our approach and DiscreteFlow recover more of the flow field than EpicFlow. On KITTI 2015, our approach is visually similar to DiscreteFlow in most street scenes (for example, scene 1). In some cases, our approach is visually more accurate (scene 2), but not on others (outliers on the white line in scene 3). Both our approach and DiscreteFlow are visually more accurate than EpicFlow.

\subsection{Controlled experiments}
\label{sec:controlled}

The generality of the presented optimization framework enables a comprehensive evaluation of different data terms and regularization terms. We perform such an evaluation using $5\%$ of the MPI Sintel training set (final pass). In all conditions, we optimize variants of the model presented in Section \ref{sec:model} using the method presented in Sections \ref{sec:optimization} and~\ref{sec:implementation}. We evaluate two data terms: the patch-based truncated NCC term given in equation (\ref{eq:NCC}) and the classical pixelwise Horn-Schunck data term given by the squared Euclidean distance in RGB color space. We also evaluate three penalty functions for the regularization term (equation \ref{eq:reg-form}): $L^1$ ($\rho(x) = |x|$), squared $L^2$ ($\rho(x) = x^2$), and Charbonnier ($\rho(x) = \sqrt{x^2+\varepsilon^2}$, where $\varepsilon=5$). For each penalty function, we evaluate a truncated regularizer ($\tau$ is determined by grid search) and a non-truncated convex form ($\tau = \infty$). All free parameters are determined by grid search.

\begin{table}[h]
\centering
\begin{tabular}{|c|c|c|}
\hline
 & Truncated & Non-truncated \\\hline
NCC+$L^1$ & {\bf 2.710} & {\bf 2.710} \\\hline
NCC+Charbonnier & 2.883 & 2.888\\\hline
NCC+$L^2$ & 2.896& 2.976\\\hline
HS+$L^1$ & 5.972 & 6.523\\\hline
HS+Charbonnier& 5.847 & 6.181\\\hline
HS+$L^2$ & 6.337 & 6.529\\\hline
\end{tabular}
\vspace{2mm}
\caption{Controlled evaluation of the data term, penalty function, and truncation. Lower is better. The patch-based NCC data term is much more effective than the pixelwise HS data term.}
\label{table:comparison}
\vspace{-3mm}
\end{table}

The results are shown in Table \ref{table:comparison}, which provides the average EPE over the images used for the evaluation for each combination of the three factors (data term, penalty function, truncation). The results suggest that the data term is of primary importance: the patch-based truncated NCC term is much more effective than the pixelwise Horn-Schunck data term, irrespective of the regularizer. Note that the non-truncated HS+$L^2$ condition corresponds to global optimization of the classical Horn-Schunck model. The results for the non-truncated NCC+$L^2$ condition indicate that by replacing the pixelwise Horn-Schunck data term with patch-based truncated NCC, retaining the classical non-truncated quadratic regularizer, and globally optimizing the objective we come within $10\%$ of the accuracy achieved by our top-performing variant. The key factors are global optimization and a patch-based data term.



\begin{figure}[h]
\centering
\includegraphics[width=0.45\textwidth]{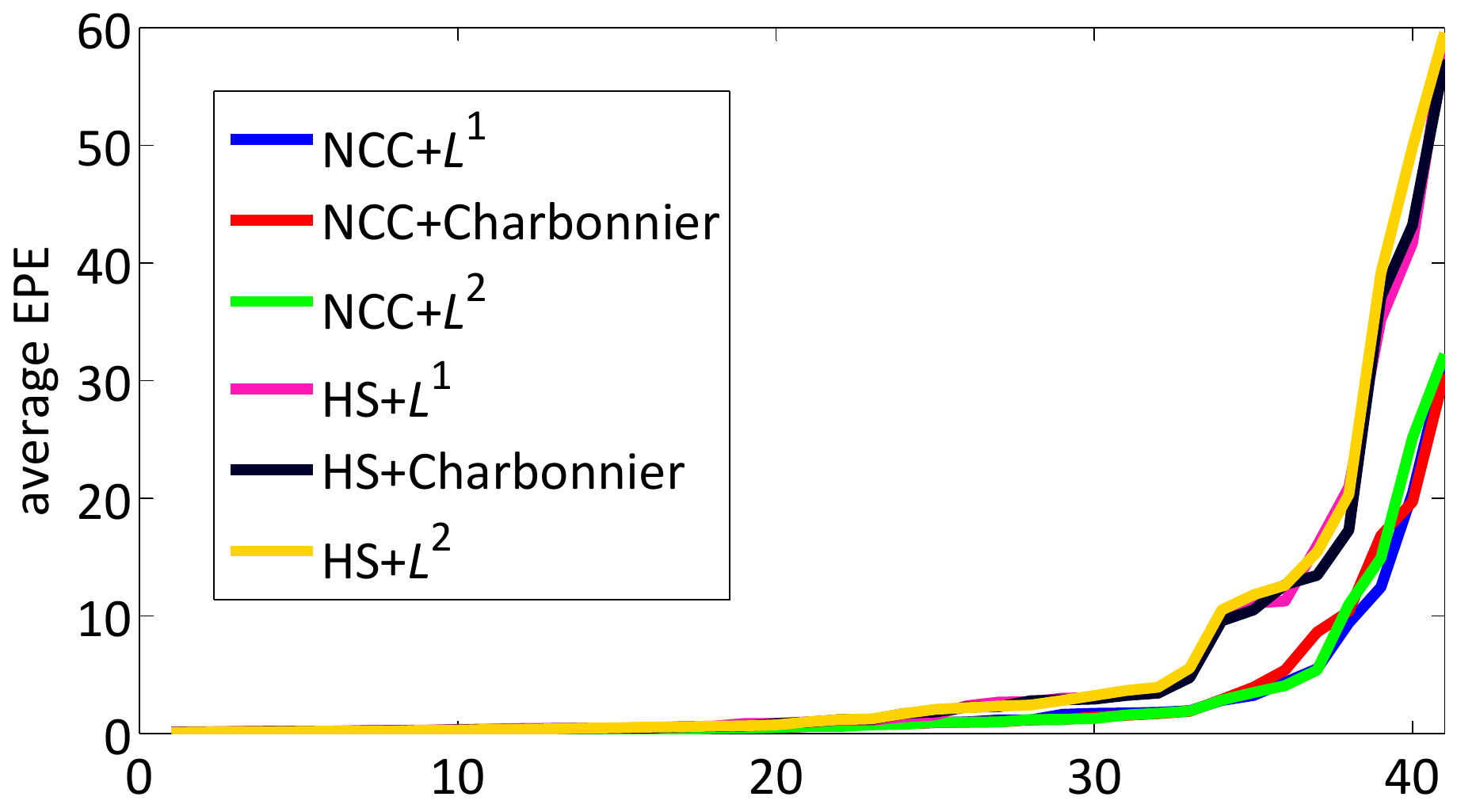}
\caption{Average endpoint error for each tested image, sorted by magnitude in each condition. Lower is better. The tested images are sorted independently for each condition: for example, image \#40 is not the same for different conditions.}
\label{figure:sorted_epe}
\end{figure}

Figure \ref{figure:sorted_epe} provides a more detailed visualization of the results. For each condition, the figure shows the average endpoint error for each tested image, sorted by magnitude. This figure shows models with truncation in the regularizer. The plots indicate that for most tested images the accuracy achieved by global optimization is high in all conditions, irrespective of the tested factors. The different conditions, specifically the two data terms, are distinguished by their robustness when accuracy is low. The patch-based NCC data term limits the error on challenging images much more effectively than the pixelwise HS data term.

To summarize, the presented optimization approach was designed to support global optimization with very general data and regularization terms. The generality of the presented framework enabled a controlled evaluation of global optimization with different data terms and regularizers. The results indicate that within a global optimization framework the detailed form of the regularizer is less important than the data term, the classical quadratic regularizer yields competitive performance, and the highest accuracy is achieved using the $L^1$ penalty.

\section{Conclusion}

We have shown that optimizing a classical Horn-Schunck-type objective globally over full regular grids is sufficient to initialize continuous interpolation and obtain state-of-the-art accuracy on challenging modern optical flow benchmarks. In particular, this demonstrates that state-of-the-art accuracy on large-displacement optical flow estimation can be achieved without externally defined descriptors. The flow objective itself is sufficiently powerful to produce accurate mappings even in the presence of large displacements. We have shown that optimizing the objective globally over the full space of mappings between regular grids is feasible and that the regular structure of the space enables significant optimizations.

Our Matlab implementation is less than 50 lines long, excluding the general-purpose TRW-S solver. We hope that the simplicity of our approach will support further advances. More advanced data terms can easily be integrated into our global optimization framework and are likely to yield even more accurate results. In addition, we believe that there is scope for further improvement in continuous interpolation accuracy, building on the impressive performance of the interpolation scheme of Revaud et al.~\cite{Revaud2015}. The output of the presented global optimization approach can serve as a canonical initialization for benchmarking such continuous interpolation schemes.



{\small
\bibliographystyle{ieee}
\bibliography{camera_ready_optical_flow}
}

\end{document}